\documentclass[draft]{agujournal2019}
\usepackage{url} 
\usepackage{lineno}
\usepackage[inline]{trackchanges} 
\usepackage{soul}
\draftfalse
\journalname{Journal of Advances in Modeling Earth Systems (JAMES)}

\usepackage[pagebackref=true,hidelinks]{hyperref}
\hypersetup{
  colorlinks = false,    
  urlcolor   = blue,    
  linkcolor  = blue,    
  citecolor  = black     
}
\hypersetup{final}
\usepackage{natbib}

\usepackage{amsmath,amsfonts,bm}

\def\eqref#1{equation~\ref{#1}}

\def\1{\bm{1}}

\def\mA{{\bm{A}}}
\def\mB{{\bm{B}}}

\def\mW{{\bm{W}}}

\DeclareMathAlphabet{\mathsfit}{\encodingdefault}{\sfdefault}{m}{sl}
\SetMathAlphabet{\mathsfit}{bold}{\encodingdefault}{\sfdefault}{bx}{n}
\newcommand{\tens}[1]{\bm{\mathsfit{#1}}}
\def\tA{{\tens{A}}}

\def\tV{{\tens{V}}}

\def\tX{{\tens{X}}}
\def\tY{{\tens{Y}}}

\def\sA{{\mathbb{A}}}

\def\sD{{\mathbb{D}}}

\def\sI{{\mathbb{I}}}
\def\sJ{{\mathbb{J}}}

\def\sM{{\mathbb{M}}}

\def\sS{{\mathbb{S}}}
\def\sT{{\mathbb{T}}}

\newcommand{\E}{\mathbb{E}}

\newcommand{\R}{\mathbb{R}}

\usepackage{url}
\usepackage{amsmath}
\usepackage{booktabs} 

\usepackage{graphicx} 
\usepackage[font=small,skip=0pt]{subcaption} 
\usepackage{balance}
\usepackage{wrapfig}
\usepackage{float} 

\usepackage[capitalize]{cleveref} 

\usepackage{amssymb} 

\usepackage[english]{babel}
\usepackage[autostyle, english = american]{csquotes}
\MakeOuterQuote{"}

\newcommand{\olive}{\textcolor{olive}}

\usepackage[normalem]{ulem} 
\usepackage[normalem]{ulem} 
\newif\ifrevision
\newif\ifstrikingthrough
\usepackage{letltxmacro}
\LetLtxMacro\origcitep\citep
\renewcommand{\citep}[2][]{
  \ifstrikingthrough
    \mbox{\origcitep[#1]{#2}}%
  \else
    \origcitep[#1]{#2}%
  \fi
}
\LetLtxMacro\origcite\cite
\renewcommand{\cite}[2][]{%
  \ifstrikingthrough
    \mbox{\origcite[#1]{#2}}%
  \else
    \origcite[#1]{#2}%
  \fi
}
\LetLtxMacro\origcitet\citet
\renewcommand{\citet}[2][]{%
  \ifstrikingthrough
    \mbox{\origcitet[#1]{#2}}%
  \else
    \origcitet[#1]{#2}%
  \fi
}
\LetLtxMacro\origcref\cref
\renewcommand{\cref}[1]{%
  \ifstrikingthrough
    \mbox{\origcref{#1}}%
  \else
    \origcref{#1}%
  \fi
}
\newcommand{\strikeolive}[1]{%
  \ifrevision
    \strikingthroughtrue
    \olive{\sout{#1}}%
    \strikingthroughfalse
  \fi
}
\newcommand{\blue}[1]{%
 \ifrevision
    {\color{blue}#1}
 \else
    {#1}%
 \fi
}
\revisionfalse 

\begin{document}

\title{The impact of internal variability on \\benchmarking deep learning climate emulators}

\authors{Björn Lütjens\affil{1}, Raffaele Ferrari\affil{1}, Duncan Watson-Parris\affil{2}, Noelle Selin\affil{1,3,4}}

\affiliation{1}{Department of Earth, Atmospheric, and Planetary Sciences, Massachusetts Institute of Technology \\ 
$^2$Scripps Institution of Oceanography and Halıcıoğlu Data Science Institute,\\University of California, San Diego\\
$^3$Institute for Data, Systems and Society, Massachusetts Institute of Technology\\
$^4$Center for Sustainability Science and Strategy, Massachusetts Institute of Technology}

\correspondingauthor{Björn Lütjens}{lutjens@mit.edu}

\justifying 
\tolerance=1600 

\begin{keypoints}
\item \blue{Linear regression outperforms deep learning for emulating 3 out of 4 spatial atmospheric variables in the ClimateBench benchmark}
\item \blue{Deep learning emulators can overfit unpredictable (multi-)decadal fluctuations, when trained on a few ensemble realizations only}
\item \blue{We recommend evaluating climate emulation techniques on large ensembles, such as the Em-MPI data subset with means over 50 realizations}

\end{keypoints}

\begin{abstract}
Full-complexity Earth system models (ESMs) are computationally very expensive, limiting their use in exploring the climate outcomes of multiple emission pathways. More efficient emulators that approximate ESMs can directly map emissions onto climate outcomes, and benchmarks are being used to evaluate their accuracy on standardized tasks and datasets. We investigate a popular benchmark in data-driven climate emulation, ClimateBench, on which deep learning-based emulators are currently achieving the best performance. \blue{We compare these deep learning emulators with a linear regression-based emulator, akin to pattern scaling, and show that it outperforms the incumbent 100M-parameter deep learning foundation model, ClimaX, on 3 out of 4 regionally-resolved climate variables, notably surface temperature and precipitation.} While emulating surface temperature is expected to be predominantly linear, this result is surprising for emulating precipitation. 
\blue{Precipitation is a much more noisy variable, and we show that deep learning emulators can overfit to internal variability noise at low frequencies, degrading their performance in comparison to a linear emulator.}
We address the issue of overfitting by increasing the number of climate simulations per emission pathway (from 3 to 50) and updating the benchmark targets with the respective ensemble averages from the MPI-ESM1.2-LR model.
Using the new targets, we show that linear pattern scaling continues to be more accurate on temperature, but can be outperformed by a deep learning-based technique for emulating precipitation. We publish our code and data at \href{https://github.com/blutjens/climate-emulator}{github.com/blutjens/climate-emulator}.
\end{abstract}

\section*{Plain Language Summary}

Running a state-of-the-art climate model for a century-long future projection can take multiple weeks on the worlds largest supercomputers. Emulators are approximations of climate models that quickly compute climate forecasts when running the full climate model is computationally too expensive. Our work examines how different emulation techniques can be compared with each other. We find that a simple linear regression-based emulator can forecast local temperatures and rainfall more accurately than a complex machine learning-based emulator on a commonly used benchmark dataset. It is surprising that linear regression is better for local rainfall, which is expected to be more accurately emulated by nonlinear techniques. We identify that noise from natural variations in climate, called internal variability, is one reason for the comparatively good performance of linear regression on local rainfall. This implies that addressing internal variability is necessary for assessing the performance of climate emulators. Thus, we assemble a benchmark dataset with reduced internal variability and, using it, show that a deep learning-based emulator can be more accurate for emulating local rainfall, while linear regression continues to be more accurate for temperature. 
\section{Introduction}
Earth system models (ESMs) are usually run many times to fully account for humanity's imperfect knowledge of future emissions and the chaotic nature of Earth's dynamics. 
But many runs come at a cost: Simulating the climate effects of a single emission pathway can take weeks and \blue{require computing resources worth hundreds of thousands of USD}
\footnote{For example, 325K USD in cloud compute assuming the reference MPI-ESM1-2-HR model~\citep{mueller18mpiesm12hr} \blue{would have been run on AWS} at 1.53 USD/36vCPU hrs with 100km res. and 5 realizations per 250-year pathway and 400 wall-clock hrs per realization on 106x 36-core nodes, \blue{totaling {$\sim$}7.6M CPU hrs}.}. 
As a result, full-complexity ESMs are often only run for a few representative emission pathways and more efficient "climate emulators" are used for applications in climate science, finance, or governance that require custom pathways~\citep{ipcc21ar6glossary,lutjens23thesis,richters24ngfs}. 
\blue{These emulators are trained to approximate the climate projections of an ESM for a given emission pathway and, recently, deep learning has been proposed as a novel technique to solve this function approximation problem. In this work, motivated by growing concerns about reproducibility and overly optimistic results in deep learning~\citep{kapoor23dataleakage}, 
we reevaluate this emerging technique from the perspective of established practices within the climate science community.} 
\strikeolive{Recently, deep learning-based climate emulators have been proposed and evaluated on benchmarks with standardized tasks and datasets~\citep{watsonparris21climatebench,kaltenborn23climateset}. However, considering a ’climate emulator’ to be any approximation of a full-complexity ESM, there exist many traditional emulation techniques, and the comparative performance of deep learning with them is still mostly unclear.}

\blue{Existing emulation techniques can be broadly distinguished into physics-based, data-driven, or hybrid combinations, following the definition in~\citep{ipcc21ar6glossary}. Physics-based techniques, such as reduced-complexity climate models, solve simplified sets of differential equations and are often focused on projecting an annually- and globally-averaged climate~\citep{meinshausen11magicc6,smith21emulators,dorheim23hectorv3}; reviewed in~\citep{sarofim21emulators}. In comparison, data-driven techniques often target higher spatial and temporal resolutions and are fitted to projections of the full-complexity ESMs~\citep{tebaldi25review}. For example, linear pattern scaling (LPS) takes global surface temperature, $\bar T$, as input and emulates the locally-resolved change of any atmospheric climate variable, such as surface temperature or precipitation, as a linear function of $\bar T$~\citep{santer90patternscaling}. LPS is possibly the most frequently used climate emulation technique, but is known to break down for emulating locally-resolved aerosol impacts, nonlinear precipitation events, or overshoot scenarios, among others~\citep{tebaldi14patternscalingstrengths,lee21ipccar6wg1patternscaling,giani24patternscaling}. Deep learning techniques may overcome these limitations, for example, by mapping regional emissions directly onto climate projections~\citep{watsonparris21climatebench}, downscaling physics-based emulators~\citep{bassetti24diffesm}, or autoregressively emulating ESM dynamics~\citep{bire23oceanfourcast,wattmeyer23ace,wang24ola}. 
With this increasing array of techniques, intercomparison projects that evaluate the advantages and disadvantages of each technique are becoming more important.}

\begin{figure}[t]
  \centering
  \subfloat{
    \includegraphics[width=0.98\linewidth]{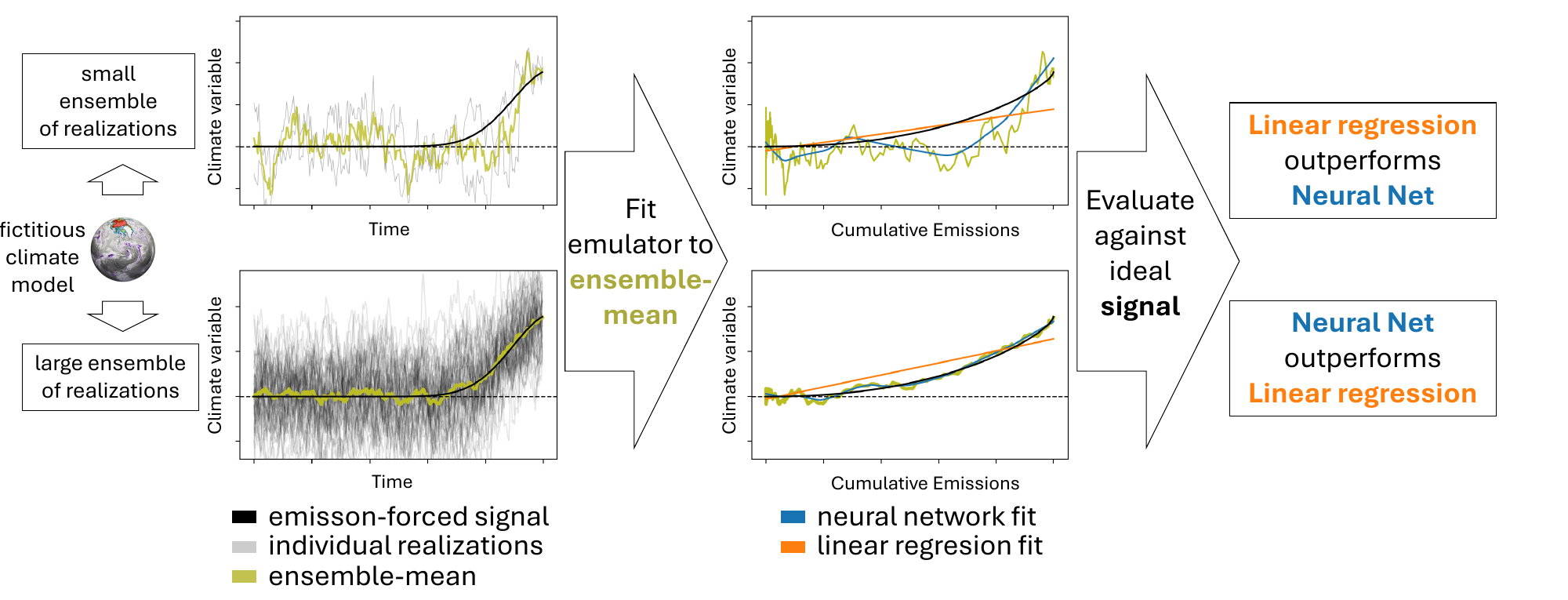}
  }
\caption[]{\blue{\textbf{A cartoon illustrating the impact of internal variability on benchmarking deep learning emulators.} The figure is generated with a mock-up stochastic model described in~\cref{sec:overfitting_exp}. The model is a nonlinear function that relates fictitious greenhouse gas emissions to nonlinear changes in an imagined climate variable. Each realization of this function is noisy, representing fluctuations from internal variability in the climate system that remains fundamentally unpredictable over long timescales.
When training an emulator to approximate this model when it has only been run a few times for the given emission scenario, neural networks can overfit to low-frequency components of these fluctuations (top-row). A simple linear approach represents the emission-forced signal more accurately, despite the opposite being the case when many realizations are available (bottom-row). To compare emulation techniques more reliably, we recommend using large climate model ensembles that attenuate the influence of internal variability.}}
\label{fig:titlefigure}
\end{figure}

\strikeolive{Existing benchmarks for deep learning-based emulators -- ClimateSet and ClimateBenchv1.0 -- do not yet include a fully representative comparison with LPS. 
Both benchmarks are designed to evaluate the performance of emulation techniques that map inputs to targets. 
Emission pathways are used as inputs and climate variables, such as locally-resolved surface temperature are used as targets. So far, only deep learning-based functions have been evaluated on the targets in ClimateSet.} \blue{Benchmarks, such as ClimateBenchv1.0, define standardized datasets, tasks, and metrics for evaluating deep learning-based emulation techniques~\citep{watsonparris21climatebench,kaltenborn23climateset}. However, they do not yet include a fully representative comparison with well-established techniques, such as LPS. 
In the ClimateBenchv1.0 benchmark (hereafter ClimateBench), emission pathways are used as inputs and climate variables, such as locally-resolved surface temperature are used as targets.} Emulators including a vision transformer (Cli-ViT), a foundation model (ClimaX), a CNN-LSTM (Neural Network), a random forest model, and a Gaussian Process have been evaluated on it, as referenced in~\cref{tab:climatebench_pattern_scaling}. LPS has also been evaluated on ClimateBench and performs very well~\citep{watsonparris21climatebench}. 
But, the existing LPS implementation used part of the target data (global mean temperatures) as inputs.
Despite this being common practice for LPS~\citep{beusch20mesmer}, it gave LPS an advantage over other emulators which were required to use only emission data as inputs. As a result, no fully representative comparison could be made and follow-up studies have cited deep-learning based emulators as the best performing techniques on ClimateBench~\citep{nguyen23climax,kaltenborn23climateset}. It remains an open question if deep learning can outperform traditional emulators.

Moreover, current climate emulation benchmarks do not sufficiently address internal variability in the climate system. Internal variability, i.e., natural fluctuations in weather or climate~(\citeauthor{leith78predictability}, 1978, and~\cref{sec:internal_var_for_ml}), can be of similar or exceeding magnitude as the emission-forced signals~\citep{hawkins09climateuncertainty}. To identify forced signals despite the presence of internal variability, researchers often take an ensemble-mean over a sufficiently large set of climate simulations with perturbed initial conditions, i.e., realizations or ensemble members~\citep{maher21smiles}. Similarly, the ClimateBench target climate variables are ensemble-means over three realizations from a single model that was run for the same emission pathways and slightly different initial conditions. However, three realizations are insufficient to separate the predictable forced signal from the unpredictable internal variability \blue{depending on the variable and timescales}, as \strikeolive{illustrated in~\cref{fig:internal_variability_graphic} and }discussed in~\citet{tebaldi21extreme} and the ClimateBench paper~\citep{watsonparris21climatebench}. \strikeolive{In ClimateSet, the core dataset for benchmarking only contains a single realization per climate model and emission pathway.}

We compare LPS with a deep learning-based emulator and investigate the impacts of internal variability on benchmarking climate emulators. 
First, we implement \blue{a representative comparison with} LPS on ClimateBench and demonstrate that LPS outperforms all deep learning-based emulators previously evaluated on ClimateBench in emulating regionally-resolved annual mean temperature, precipitation, and extreme precipitation (according to the ClimateBench evaluation protocol). This good performance of LPS is notable given the known nonlinear relationships in precipitation. \strikeolive{illustrated in~\cref{fig:linearity_s_e_asia}} 
Then, we use data from the MPI-ESM1.2-LR ensemble with 50 realizations per emission pathway~\citep{olonscheck23mpismile}, hereafter abbreviated as \textit{Em-MPI data},
to evaluate the effect of internal variability on benchmarking. \blue{As illustrated for an idealized problem in~\cref{fig:titlefigure},} we demonstrate that high levels of internal variability in the target data can make deep learning emulators prone to overfitting and skew the benchmarking results in favor of emulation techniques with lower model complexity, such as LPS. After reducing the internal variability by averaging over many realizations, we find that a deep learning-based emulator (CNN-LSTM) can be more accurate than LPS for spatial precipitation, while LPS continues to be the more accurate emulation technique for surface temperature. In summary, addressing internal variability is necessary to make conclusive comparisons between emulation techniques.

\section{Data \& Methods}\label{sec:data}
In~\cref{sec:climatebench_data}, we review the targets, inputs, evaluation protocol, and metrics that are used in ClimateBench. In~\cref{sec:mpi_data}, we describe the Em-MPI data and the associated update to the targets. In~\cref{sec:lps_model}, we detail LPS and, in~\cref{sec:cnn_lstm_review}, our reimplementation of the CNN-LSTM from~\citep{watsonparris21climatebench} with specifics in~\cref{app:cnn-lstm}.
In~\cref{sec:internal_var_exp}, we introduce our ``internal variability experiment'' which uses the Em-MPI targets to measure how internal variability affects the relative performance of these two emulators. \blue{In~\cref{sec:overfitting_exp}, we introduce a simplified dataset to test if deep learning-based models can overfit to internal variability.}

\subsection{Background on the ClimateBench benchmark}\label{sec:benchmark_data}\label{sec:climatebench_data}

\blue{Key components of an ML benchmark are the input and target datasets and an evaluation protocol that prescribes data splits and evaluation metrics~\citep{dueben22benchmarks}.} The ClimateBench inputs and targets are derived from emission pathways and associated climate projections from three realizations of the climate model, NorESM2-LM~\citep{seland20noresmlm}, for each of seven emission scenarios, $e\in\sS_e=$ \{hist-ghg, hist-aer, historical, ssp126, ssp245, ssp370, ssp585\}.
The ``ssp'' and ``hist-'' scenarios are Shared Socioeconomic Pathway- (ssp) and single-forcer-experiments in the ScenarioMIP and DAMIP activities within the Coupled Model Intercomparison Project Phase 6 (CMIP6)~\citep{oneill16ssp,gillett16damip,eyring16cmip}. The data for each historical scenario encapsulates $T_\text{hist}=165$ (1850-2014) years and $T_\text{spp}=86$ (2015-2100) years for each ssp, and, each realization is distinguished by a different initial condition. 

The ClimateBench targets are locally-resolved annually-averaged climate variables, as projected by the climate model. These target variables are surface temperature ($\mathrm{tas}$), diurnal temperature range ($\mathrm{dtr}$), precipitation ($\mathrm{pr}$), and 90th percentile precipitation ($\mathrm{pr90}$). All variables are ensemble-mean anomalies, i.e., during preprocessing the projections were averaged over three realizations per emission scenario and long-term averages of a 500-year (1350-1850) preindustrial control run were subtracted from it. 
The dimensionality of the target dataset of each variable is $I\times J\times \sum_{e\in\sS_e} T_e$ with $I=96$ and $J = 144$ latitudinal and longitudinal grid points corresponding to a ($1.875^\circ\times 2.5^\circ$) model resolution and $T_e = \{T_\text{hist}$ if $e$ starts with $hist$; $T_\text{ssp}$ if $e$ starts with $ssp$\} simulated years.


The ClimateBench input dataset contains four variables for each emission scenario: globally-averaged cumulative emissions of carbon dioxide ($\mathrm{CO}_2^\text{cum}$), globally-averaged emissions of methane ($\mathrm{CH}_4$), and spatial maps of emissions of sulfur dioxide ($\mathrm{SO}_2$) and black carbon ($\mathrm{BC}$). For each year, $\mathrm{BC}$ and $\mathrm{SO}_2$ are locally-resolved ($I\times J$) matrices and $\mathrm{CO}_2^\text{cum}$ and $\mathrm{CH}_4$ are globally-aggregated scalars. The variables are detailed in~\cref{app:climatebench}.

The ClimateBench evaluation protocol prescribes the following data splits and evaluation metrics. 
All scenarios except for \strikeolive{ssp245} \blue{one} are used as the training set, i.e., $\sS_{e,\text{train}} = \{\text{hist-ghg, hist-aer, historical, ssp126, ssp370, ssp585}\}$, resulting in $\sum_{e\in\sS_{e,\text{train}}} T_e=753$ training years that are used by an optimizer to find the best model weights or parameters. \blue{The ClimateBench protocol does not prescribe a separate validation set, meaning that emulation techniques may hold out any samples from the training set and use them as validation set for choosing hyperparameters.}
The years 2080-2100 in ssp245 are used as the evaluation or test set, which is recommended to be unseen until reporting the scores in the results table~\citep{goodfellow16deeplearning}. The years 2015-2079 in ssp245 are not used due to their similarity to training and test scenarios. 

The evaluation metrics in the ClimateBench results table in~\cref{tab:climatebench_pattern_scaling} are "spatial", "global", and "total" normalized root mean square errors (NRMSEs) between the predicted, $\hat \tY_e$, and ensemble-mean target climate variable, $\mathbb{E}_{\sM}[\tY_e]$. We provide equations for the normalization coefficient and "total" $\mathrm{NRMSE}$ in~\cref{app:climatebench_metrics}. The spatial and global scores, $\mathrm{RMSE}_s$ and $\mathrm{RMSE}_g$, which are used in the ClimateBench NRMSE and our experiments are calculated as follows:

\begin{equation}
    \mathrm{RMSE}_s(\hat \tY_e, \mathbb{E}_{\sM}[\tY_e]) = 
    \frac{1}{\sum_i \alpha_i J} \sum_{i,j}\alpha_i
    \sqrt{
    \left(\frac{1}{\lvert\sT_\text{test}\rvert}\sum_{t} \hat y_{i,j,t,e} - \frac{1}{\lvert\sT_\text{test}\rvert}\sum_{t} y_{i,j,t,e}\right)^2}
    \label{eq:rmse_spatial}
\end{equation}

\begin{equation}
    \mathrm{RMSE}_g(\hat \tY_e, \mathbb{E}_{\sM}[\tY_e]) = 
    \frac{1}{\lvert\sT_\text{test}\rvert}\sum_{t}
    \sqrt{
    \left(\frac{1}{\sum_i \alpha_i J}\sum_{i,j}\alpha_i\hat y_{i,j,t,e} - \frac{1}{\sum_i\alpha_i J}\sum_{i,j}\alpha_i y_{i,j,t,e}\right)^2}
    \label{eq:rmse_global}
\end{equation}
where the $i,j,t,m,e$ indices refer to latitude, longitude, year, realization ID, and emission scenario, respectively. First, the ensemble-mean over all realizations is computed, $\mathbb{E}_{\sM}[\tY_e] = \frac{1}{\lvert\sM\rvert} \sum_{m} \tY_{m,e}$ with $y_{i,j,t,e} = \frac{1}{\lvert\sM\rvert}\sum_m y_{i,j,t,m,e}$. Then, the spatial RMSE computes the per-pixel error of 21-year averages (2080-2100). The global RMSE computes the per-year error of global averages. Global averages are taken with latitude weights, $\alpha_i = \cos(\phi_i)$, where $\phi_i\in[-\frac{\pi}{2}, \frac{\pi}{2}]$. The indices are in the sets $i\in\sI =\{1,2,...,I\}$, $j\in\sJ =\{1,2,...,J\}$, $t\in\sT_\text{test} =\{2080,2081,...,2100\}$, $m\in\sM =\{1,2,N\}$, and $e\in \sS_{e,\text{test}}=$\{ssp245\}. The ClimateBench dataset has $N=3$ realizations. The sums over those sets are abbreviated as $\sum_{\text{index}\in\text{set}} =: \sum_{\text{index}}$. We summarize the climate variables by denoting the tensors $\tY_e = \{y_{i,j,t,m,e}\}_{i\in\sI, j\in\sJ, t\in\sT_\text{test}, m\in\sM}$ and $\hat \tY_e = \{\hat y_{i,j,t,e}\}_{i\in\sI, j\in\sJ t\in\sT_\text{test}}$. 
In general, we denote a tensor as $\tA$, matrix as $\mA$, set as $\sA$, cardinality of a set as $\lvert\sA\rvert$, and scalar as $a$. 

\begin{figure}[t]
  \centering
  \subfloat[Surface temperature anomalies]{
    \includegraphics[width=0.97\linewidth]{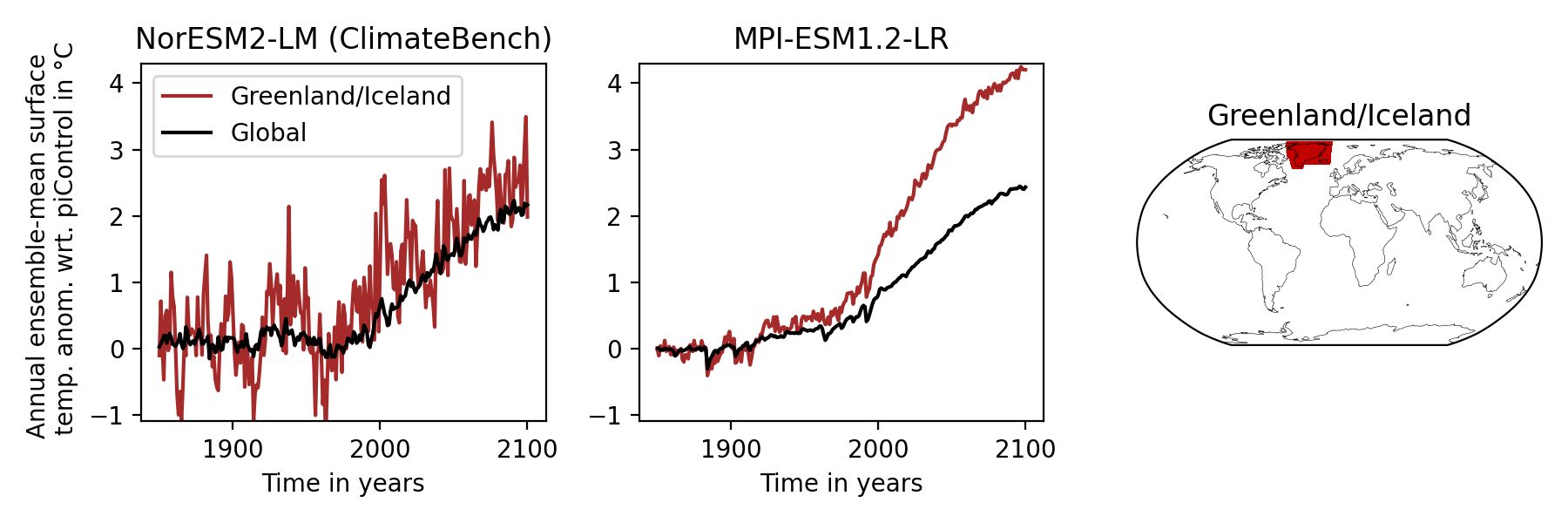}
    \label{fig:internal_variability_tas}
    \vspace{-0.05in}
  }\\
  \subfloat[Precipitation anomalies]{
    \includegraphics[width=0.97\linewidth]{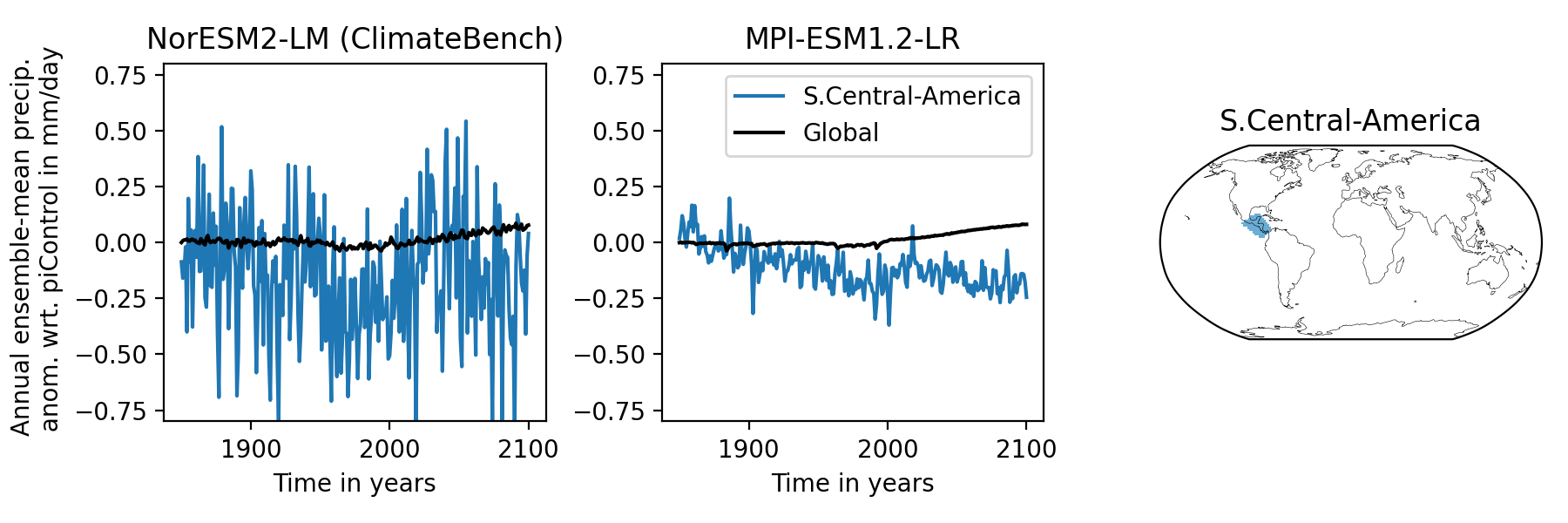}
    \label{fig:internal_variability_pr}
    \vspace{-0.05in}
  }
\caption[Internal variability]{\textbf{Internal variability in 3-member NorESM2-LM vs. 50-member MPI-ESM1.2-LR ensemble-mean}. The plots show the ensemble-mean of the global (black) and regionally-averaged surface temperature (top-red) and precipitation (bottom-blue) anomalies from the $historical$ and $ssp245$ scenario. The left plots show the NorESM data that is used in the ClimateBench test set and the middle plots show the Em-MPI data. The regional averages are calculated over displayed IPCC AR6 reference regions (right). The interannual fluctuations from internal variability are visibly smaller in the 50-member Em-MPI data in comparison to the 3-member NorESM data, especially at the regional scale. More regions are shown in~\cref{fig:internal_variability_selected_regions}.}
\label{fig:internal_variability_graphic}
\vspace{-0.2in}
\end{figure}

\subsection{Em-MPI data: Addressing internal variability with the MPI-ESM1.2-LR ensemble}\label{sec:mpi_data}

We select the MPI-ESM1.2-LR (v1.2.01p7)~\citep{mauritsen19mpiesm1-2-lr} single model initial-condition large ensemble~\citep{olonscheck23mpismile} for extracting the data subset, \textit{Em-MPI}. We choose the MPI model instead of other climate models, by first subselecting the CMIP6 models from the multi-model large ensemble archive, MMLEAv1 by~\citet{deser20mmlea}, that ran at least 30 realizations \blue{by 2022/12/01} for the Tier 1 scenarios in ScenarioMIP (ssp126, -245, -370, and -585). This leaves only three models: MPI-ESM1.2-LR, EC-Earth3, and CanESM5. \blue{
We choose the MPI model for its transient climate response within the ``likely'' range and near-immediate availability of all realizations, though other reasonable choices exist (see~\cref{app:em_mpi}).} \strikeolive{Out of this selection, the MPI model is the only one with a transient climate response within the ``likely'' range and a sufficient number of online-available realizations, as detailed in~\cref{app:em_mpi}.} Further, MPI-ESM1.2-LR simulates internal variability in surface temperature comparable to observations and only slightly underestimates the variability in precipitation~\citep{olonscheck23mpismile}.

The Em-MPI dataset contains 50 realizations for each emission scenario, which can be used to average out or reduce the interannual fluctuations from internal variability and extract the emission-forced signal as illustrated in~\cref{fig:internal_variability_graphic} and~\cref{sec:em_mpi_internal_var}.
We download the 50 realizations for annual mean surface temperature and precipitation for the scenarios $\mathbb{S}_e^\text{em-mpi}=$ \{ssp126, ssp245, ssp370, ssp585, historical\} at T63 horizontal resolution ($\approx$ 1.8° or 180km; $96\times192$ pixels). 
We do not regrid this dataset. Each ssp and historical scenario contains $T_\text{ssp}=86$ or $T_\text{hist}=165$ years, respectively. \blue{We compute anomalies by taking each annually-averaged variable and subtracting the preindustrial climatology, i.e., the locally-resolved average over a 400-year preindustrial control run during year 1450 to 1850.}

As hourly precipitation rates are approximately log-normally distributed~\citep{hyekyung04lognormalprecip}, it might be desirable to download hourly instead of annually-averaged precipitation and take a log transform of precipitation before computing ensemble and annual averages. 
However, we maintain precipitation in its original scale to keep results comparable between the Em-MPI and NorESM2-LM targets and because annually-averaged precipitation is a well studied quantity. Furthermore, we do not take the log of annually-averaged precipitation anomalies because, consistent to the central limit theorem, their distribution more closely resembles a Gaussian rather than a log-normal (\cref{fig:mpi_distribution_pr} shows the actual distribution).



\subsection{Linear Pattern Scaling emulator}\label{sec:lps_model}
\begin{figure}[t]
  \centering
  \subfloat{
    \includegraphics[width=0.98\linewidth]{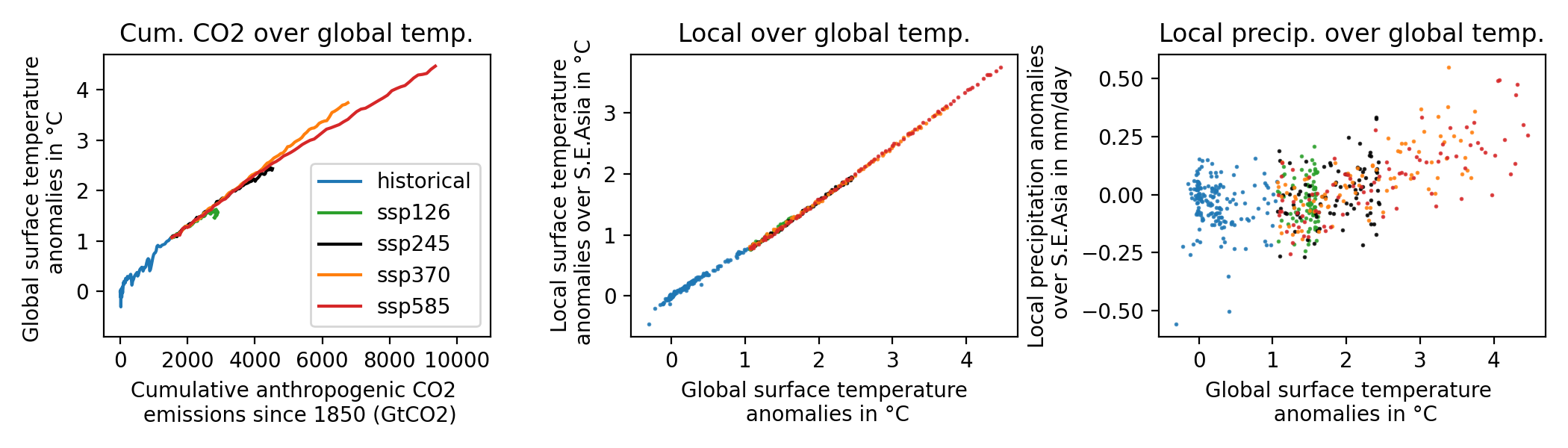}
  }
\caption[]{\textbf{Functional relationships in cumulative $\mathrm{CO_2}$ emissions, surface temperature, and precipitation.} The left plot shows the almost linear relationship between cumulative $\mathrm{CO}_2$ emissions (x-axis) and global ensemble-mean surface temperature anomalies (y-axis) for each scenario in the Em-MPI data. 
The middle and right plot show the ensemble-mean local surface temperature and precipitation anomalies, respectively, averaged over each year (dot) and the IPCC AR6 region S.E.Asia, against the annually-averaged global surface temperature anomaly.  
We selected S.E.Asia to highlight the contrast between linear and more complex relationships in temperature vs. precipitation and show other regions in~\cref{fig:linearity_for_many_regions}. 
}
\label{fig:linearity_s_e_asia}
\end{figure}

Linear Pattern Scaling (LPS) is a commonly used emulation technique to analyse regional climate change as a function of global emissions~\citep{mitchell99patternscaling, giorgetta13mpigepatterns}\blue{; see review by~\citet{lee21ipccar6wg1patternscaling}}. In our LPS emulator, $\hat f_{\text{LPS}}$, a linear regression fit first relates scalar global cumulative $\mathrm{CO}_2$ emissions, $x_{t,e} = \mathrm{CO}_{2;t,e}^{cum} \in\R$, to global surface temperature anomalies, $\overline{T}^{\text{surf}}_{t,e}\in\R$, for a chosen year, $t\in\{1850, 1851, ..., 2100\}$, and emission scenario, $e\in \mathbb{S}_e$:

\begin{equation}
\hat{\overline{T}}^{\text{surf}}_{t,e} = w^{\text{global}} x_{t,e} + b^{\text{global}}.
 \label{eq:lps_global}
\end{equation}

This regression for global temperature does not utilize other emission variables consistent with the evidence that $\overline{T}^{\text{surf}}_{t,e}$ is predominantly linearly proportional to $\mathrm{CO}_{2;t,e}^\text{cum}$ (see~\citep{macdougall16tcre} and~\cref{fig:linearity_s_e_asia}-left). Many techniques emulate the global mean temperature more accurately than a linear fit, for example, by considering non-$\mathrm{CO}_2$ forcings or multiple time-scales~\citep{meinshausen11magicc6,smith18fairv13}, but we are choosing to sacrifice accuracy for added simplicity and interpretability. 

Then, we fit local linear regression functions independently for each grid cell, $(i,j)$, to map global temperature anomalies onto the local climate variable of interest, $y_{i,j,t,e}\in\R$: 

\begin{equation}
 \hat{y}_{i,j,t,e} = w^{\text{local}}_{i,j} \hat{\overline{T}}^{\text{surf}}_{t,e} + b^{\text{local}}_{i,j}
\label{eq:lps_local}
\end{equation}

This regression assumes that any local climate variable can be described as a linear function of the global mean temperature and that this function is consistent across emission scenarios. This assumption is known to be relatively accurate for local surface temperature and less accurate for local precipitation (see~\citep{tebaldi14patternscalingstrengths} and ~\cref{fig:linearity_s_e_asia}-middle and -right). The existing implementation of LPS on ClimateBench~\citep{watsonparris21climatebench} uses $\overline{T}^{\text{surf}}_{t,e}$ from the target dataset as inputs (instead of estimating the variable from emissions). Our LPS emulator can be represented as matrix operations: $\hat{\tY}_{t,e} = \hat f_\text{LPS}(x_{t,e}) = \mW^{\text{local}} (w^{\text{global}} x_{t,e} + b^{\text{global}}) + \mB^{\text{local}}$ with $\hat \tY_{t,e} = \{\hat y_{i,j,t,e}\}_{i\in\sI,j\in\sJ}$. The LPS emulator including the global and local weights and intercepts, $w^{\text{global}}, b^{\text{global}}, w^{\text{local}}_{i,j}, b^{\text{local}}_{i,j}\in\R$, are independent of time and scenario. We split the global and local regression into~\cref{eq:lps_global,eq:lps_local} for interpretability, but adding a global weight and intercept is redundant; thus, the total number of free parameters is the number of local weights and intercepts: $2*I*J$. By construction, LPS preserves the global average if the local \blue{variable} and the global variable are \blue{both} temperature (Sect.2.2 in \citet{wells23patternscaling}). The optimal weights are found via ordinary least squares with the Python package scikit-learn~\citep{pedregosa11sklearn}.

\subsection{Review of the CNN-LSTM emulator}\label{sec:cnn_lstm_review}
A CNN-LSTM is a commonly used neural network architecture for learning spatiotemporal correlations~\citep{shi15cnnlstm} and has been evaluated by~\citet{watsonparris21climatebench} on ClimateBench. In comparison to LPS, the CNN-LSTM is expected to be capable of learning more complex functions by having more free parameters. The CNN-LSTM also contains a 10-year memory and incorporates all emission variables, including spatially resolved aerosols, as inputs.

\blue{To increase the comparability between the CNN-LSTM scores on the Em-MPI and NorESM data, we follow the implementation in the ClimateBench paper as closely as possible. This includes using the same hyperparameters, because the reference implementation did not withhold any samples for validation and did not go through the process of hyperparameter optimization.}
\strikeolive{We follow the implementation in~\citep{watsonparris21climatebench} to reimplement the CNN-LSTM for our internal variability experiment,} However, we noticed that the CNN-LSTM converges to significantly different minima depending on the weight initialization random seed. To compensate for this spurious dependence on initialization, we deviate from the original implementation by training the CNN-LSTM multiple times with different weight initialization random seeds and reporting the average scores across those seeds whenever we evaluate the CNN-LSTM. \blue{We also terminate the CNN-LSTM training process when the test error stops improving to reduce computational cost and avoid memorization of the training data.} See~\cref{app:cnn-lstm} for details on this CNN-LSTM.

\subsection{Testing the influence of internal variability on emulator performance}\label{sec:internal_var_exp}

We construct the following experiment to quantify the impact of internal variability on emulator performance (with results in~\cref{fig:spatial_pr_over_realizations} and~\cref{sec:results_internal_variability}).
We fit two emulation techniques (LPS and CNN-LSTM) on random ensemble subsets of $n$ realizations and evaluate them for each $n$ on the mean of all $50$ realizations. We evaluate on the mean of $50$ (instead of $n$) realizations, because we aim to benchmark emulators of the forced mean climate response~\citep{maher19mpige}. We assume that $50$ realizations are sufficient to approximate the true forced climate response of a given climate model, which is a reasonable assumption for most variables and spatiotemporal resolutions~\citep{tebaldi21extreme,olonscheck23mpismile}.
Thus, evaluating against the mean of $50$ realizations will determine for each $n$-member training set which technique we define as the "better" emulator for the MPI-ESM1.2-LR model. We use `member' and `realization' interchangeably.


In detail, we randomly draw the ensemble subsets, $\sM_{n,k} \subset \sM = \{1,2,...,N^\text{em-mpi}\}$, where $n=\lvert \sM_{n,k}\rvert$ denotes the number of realizations in a subset, $k$ indexes the $k$th random draw of equally sized subsets, and $m\in\sM_{n,k}$ is the realization ID. Each ensemble subset is drawn from the Em-MPI data that has $N^\text{em-mpi}=50$ realizations without replacement, i.e., within each subset there are no duplicate realization IDs. To compensate for the variation across subsets, we draw $K=20$ ensemble subsets of the same size for every $n$ and report the mean and standard deviation. The subsets, for example, contain the realization IDs, $\sM_{1,1}=\{3\}, \sM_{1,2}=\{48\}, ..., \sM_{3,1}=\{2,18,36\},\sM_{3,2}=\{1,10,42\}, $ etc.

We fit each emulation function, $\hat f_\text{LPS}$ and $\hat f_\text{CNN-LSTM}$, on the ensemble-mean of each ensemble subset, $\mathbb{E}_{\sM_{n,k}} [\tY_\text{train}] = \frac{1}{\lvert\sM_{n,k}\rvert}\sum_{m} \tY_{m,\text{train}}$, and denote the fitting or training process with a $P$. 
We use the same inputs as in ClimateBench, targets from the Em-MPI data, and a \blue{train/test} data split equivalent to the one in ClimateBench, i.e., $\sS_{e,\text{train}}^\text{em-mpi}=$\strikeolive{$\sS_{e,\text{val}}^\text{em-mpi}=$}\{historical, ssp126, ssp370, ssp585\} and $\sS_{e,\text{test}}^\text{em-mpi}=$\{ssp245\} with $\sT_\text{test}=\{2080,2081,...,2100\}$, $\tX_\text{train} = \{\tX_e\}_{e\in\sS_{e,\text{train}}^{\text{em-mpi}}}$, and $\tY_\text{train} = \{\tY_e\}_{e\in\sS_{e,\text{train}}^{\text{em-mpi}}}$. After training, we evaluate the emulators on the ensemble-mean of all $50$ realizations, $\mathbb{E}_\sM[\tY_\text{test}]$.
As evaluation metrics, we use the spatial and global $\mathrm{RMSE}_*$ (from~\cref{eq:rmse_spatial,eq:rmse_global}), averaged over the ensemble subsets, and the difference between the emulator scores, $\Delta \mathrm{RMSE}_{*,n} = \mathrm{RMSE}_{*,n,\text{CNN-LSTM}} - \mathrm{RMSE}_{*,n,\text{LPS}}$. The train and evaluation process is conducted independently for precipitation and surface temperature. In summary:

\begin{align}
    \hat f_{n,k} &\xleftarrow{\text{fit}} P\left[\tX_\text{train},  \mathbb E_{\sM_{n,k}}[ \tY_{\text{train}}]\right]\\
    \mathrm{RMSE}_{*,n} &= \frac{1}{K}\sum_{k\in\{1,...,K\} }\mathrm{RMSE}_*\left(
    \hat f_{n,k}\left(\tX_\text{test} \right),
    \mathbb{E}_{\sM} [\tY_{\text{test}}]
    \right)
\end{align}

We use the LPS and CNN-LSTM as described in~\cref{sec:lps_model,sec:cnn_lstm_review}, respectively. We train the CNN-LSTM $L=20$ times with different weight initialization random seeds per realization subset draw and use the average RMSEs of those $20$ seeds as the scores for a given subset draw, as detailed in~\cref{eq:rmse_cnn}. We fit the LPS emulator for every $n\in\{1,2,...,50\}$ which results in $K*50=1000$ runs and the CNN-LSTM for $n\in\{1,2,...,15,16,20,25,30,40,50\}$ resulting in $L*K*21=8400$ runs.

\blue{We report the spatial and global RMSE for training on the 50-member ensemble-mean targets in~\cref{tab:em_mpi_scoreboard}. We computed scores for LPS and the CNN-LSTM reimplementation, and we recommend scientists that add emulation techniques to clarify that ``scores are computed on the Em-MPI targets using the ClimateBench evaluation protocol''.}

\subsection{A heuristic model to illustrate internal variability effects}\label{sec:overfitting_exp}

\blue{
To better understand why internal variability influences the emulator performance, we repeat the internal variability experiment on a simplified dataset. We generate the dataset with a one-dimensional nonlinear model that relates emissions to a hypothetical climate variable with internal variability. The model is based on an energy balance budget (e.g., \citeauthor{giani24patternscaling} \citeyear{giani24patternscaling}) that simulates temperature change as a function of cumulative emissions of a fictitious greenhouse gas, climate feedbacks, and noise. The noise is sampled from a Wiener process, representing internal variability generated by chaotic atmospheric dynamics~\citep{hasselmann76ouprocess}. 
Integrating the noise over time creates unpredictable high- and low-frequency fluctuations, comparable to interannual and (multi-)decadal oscillations~\citep{frankignoul77stochastic}. The hypothetical climate variable is an arbitrary nonlinear function of temperature, comparable to precipitation in the real climate system. 

We use the ensemble-mean across $n$ independent realizations of this model as one training set, and repeat this to generate $K=2000$ training sets for every $n$.
We fit a neural network and a linear regression function to each training set and, in comparison to~\cref{sec:internal_var_exp}, train the fixed-complexity neural networks by following best practices to avoid overfitting, such as using a separate validation set and tuning regularization parameters. After training, we evaluate both emulation techniques using the expected MSE for every $n$ with respect to the noise-free emission-forced signal. We detail the heuristic model and training process in~\cref{app:ou_experiment_setup}.
}

\blue{
Following the bias-variance decomposition, we decompose the MSE scores into a bias-squared and variance term, $\text{MSE}_n = \text{Bias}_n^2 + \text{Var}_n$, as proven in~\cref{app:bias_variance_decomposition}. This decomposition allows us to interpret each emulation technique in the context of the well-studied bias-variance tradeoff~\citep{bishop06bishop}. We also compute the expected Fourier spectra of the neural network and linear fits to identify if the methods are influenced by lower or higher frequency noise components.}

\section{Results}

In~\cref{sec:results_climatebench}, we quantify the performance of LPS on ClimateBench. For spatial precipitation, LPS achieves better scores than deep learning-based emulators even though the relationships are known to be nonlinear (discussed in~\cref{sec:lps_limitations}). 
In~\cref{sec:internal_var}, we illustrate that the internal variability in the 3-member ClimateBench targets is high in comparison to the 50-member Em-MPI data. Then, we illustrate that this high internal variability can skew benchmarking scores in favor of LPS, i.e., LPS often has comparatively better scores than the CNN-LSTM if there is more internal variability in the training set. Lastly, we explain this effect of internal variability by demonstrating the associated bias-variance tradeoff on a simplified dataset.

\subsection{Evaluation of Linear Pattern Scaling on ClimateBench}\label{sec:results_climatebench}
\begin{figure}[t]
  \centering     
  \subfloat{
    \includegraphics[width=0.98\linewidth]{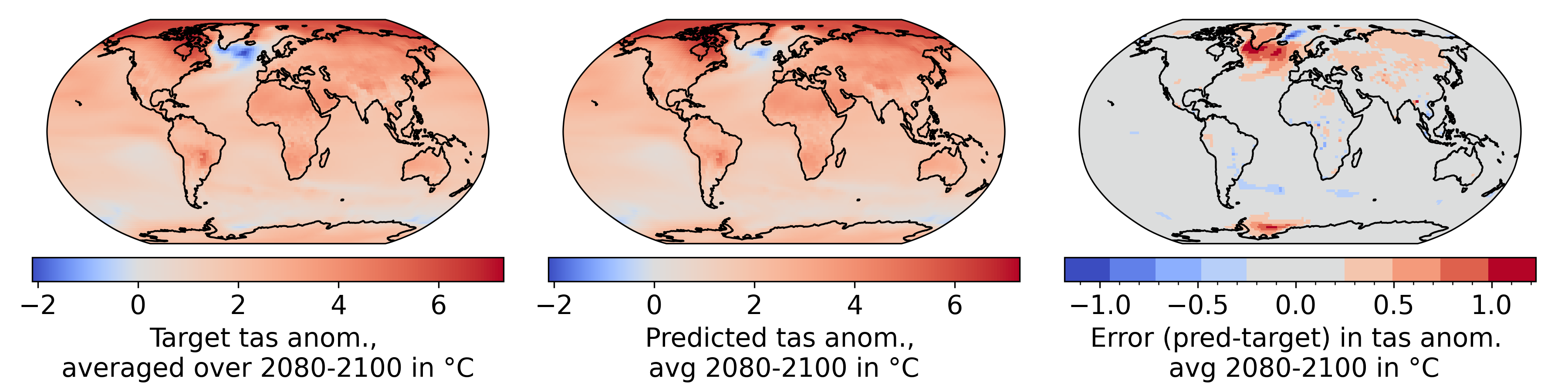}
  }
\caption[Linear pattern scaling error map tas ssp245]{\textbf{Linear pattern scaling error map}. The left plot shows the target surface temperature anomalies ($\mathrm{tas}$) from the ssp245 ClimateBench test set, which are averages over 3 realizations and 21 years (2080-2100). The middle plot shows the linear pattern scaling predictions and the right plot the error of predictions minus the target. The other variables are plotted in~\Cref{fig:linear_pattern_scaling_error_map_diurnal_temperature_range,fig:linear_pattern_scaling_error_map_precipitation,fig:linear_pattern_scaling_error_map_90th_precipitation}.
}
\label{fig:linear_pattern_scaling_tas_ssp245}
\end{figure}

\subsubsection{Regional accuracy of LPS}\label{sec:results_lps_regional_acc}

~\Cref{fig:linear_pattern_scaling_tas_ssp245} shows that LPS (middle) predicts surface temperature patterns that are similar to the ClimateBench test set targets (left). Important warming patterns are visible in the LPS predictions; for example, a North Atlantic "cold blob", Arctic amplification, and proportionately higher warming over land and the Northern Hemisphere.~\Cref{fig:linear_pattern_scaling_tas_ssp245}-right shows that LPS predicts the surface temperature anomalies in most regions within a $\pm 0.25^\circ C$ error bar of the target ESM-simulated values. The global patterns in the LPS predictions also resemble the target patterns for other climate variables, as shown in~\cref{fig:linear_pattern_scaling_error_map_diurnal_temperature_range,fig:linear_pattern_scaling_error_map_precipitation,fig:linear_pattern_scaling_error_map_90th_precipitation}. Limitations of LPS are visible in regional errors, for example, in the surface temperature at the North Atlantic cold blob and Weddell Sea. These errors could arise from nonlinear relationships or internal variability, as discussed in existing LPS evaluations~\citep{wells23patternscaling} and~\cref{sec:lps_limitations}.

\subsubsection{Quantitative comparison of LPS with deep learning emulators}
\Cref{tab:climatebench_pattern_scaling} compares LPS against all other emulators that have reported scores on ClimateBench including the transformer-based foundation model, ClimaX, with 108M free parameters that is currently being cited as the best performing emulator~\citep{kaltenborn23climateset}. The Gaussian Process and Random Forest entries are described in~\citep{watsonparris21climatebench}. 
The CNN-LSTM by~\citep{watsonparris21climatebench} is summarized in~\cref{app:cnn-lstm}, CNN-LSTM (reproduced) is reimplementation of this CNN-LSTM by~\citep{nguyen23climax}, and Cli-ViT is a vision transformer detailed in~\citep{nguyen23climax}. We highlight the emulator with the lowest NRMSE score (i.e. best performance) in \textbf{bold}. But, we caution the reader not to draw a definitive conclusion which emulation technique is most accurate beyond the ClimateBench benchmark because of the internal variability in the NorESM2-LM targets, as analysed in~\cref{sec:internal_var}. 

LPS has the lowest spatial NRMSE scores for 3 out of 4 variables including surface temperature, precipitation, and 90th percentile precipitation (see ~\cref{tab:climatebench_pattern_scaling}).
Further, LPS requires 4 orders of magnitude fewer parameters than ClimaX. This result implies that LPS is the new incumbent on spatial variables, i.e., the emulation technique that achieves the lowest score in the most "Spatial" columns, on ClimateBench.


ClimaX and the CNN-LSTM (reproduced) have better scores than LPS on diurnal temperature range, which is strongly influenced by aerosols, as discussed in~\citep{watsonparris21climatebench}. ClimaX and CNN-LSTM (reproduced) also have better global scores in precipitation and 90th percentile precipitation which we discuss in~\cref{app:climatebench_pattern_scaling}. Giving every column in the results table equal weight, LPS and ClimaX are on par with each other with each having better scores on 6 out of 12 columns.


LPS outperforming deep learning on surface temperature aligns with the domain intuition that there is a linear relationship to cumulative $\mathrm{CO}_2$ emissions. But, it is surprising that LPS also outperforms the nonlinear deep learning-based emulators on spatial precipitation even though the relationships are known to be nonlinear. 

\begin{table}
\caption[ClimateBench results table]{\textbf{ClimateBench results table including linear pattern scaling}. Spatial, Global, and Total denote normalized root mean square errors (NRMSEs) as detailed in~\cref{eq:rmse_spatial,eq:rmse_global,eq:nrmse_spatial}.
The standard deviation of the three target ensemble members is denoted as Target std. dev. (WP22). We use three significant digits unless reported otherwise in the reference($\prime$). The references are WP22~\citep{watsonparris21climatebench} and N23~\citep{nguyen23climax}.}
\label{tab:climatebench_pattern_scaling}
\centering
\resizebox{0.99\textwidth}{!}{
\begin{tabular}{l llclllllllllll}
\toprule
 &   ref.&$\#$param&\multicolumn{3}{c}{Surface temperature} &   \multicolumn{3}{c}{Diurnal temperature range}&\multicolumn{3}{c}{Precipitation} & \multicolumn{3}{c}{90$^\text{th}$ percentile precipitation}\\
\cmidrule(r){2-2}\cmidrule(r){3-3}\cmidrule(r){4-6}\cmidrule(r){7-9}\cmidrule(r){10-12}\cmidrule(r){13-15}
&   &&Spatial& Global&Total &   Spatial&Global&  Total&Spatial&Global&Total & Spatial& Global&Total\\
\midrule
Gaussian process&   WP22& n/a &0.109
& 0.074'& 0.478
&   9.21&2.68&  22.6
&2.34&0.341&4.05
 & 2.56& 0.429&4.70
\\
CNN-LSTM&   WP22&365K&0.107
& 0.044'& 0.327
&   9.92&1.38&  16.8
&2.13&0.209&3.18
 & 2.61& 0.346&4.34
\\
CNN-LSTM (reproduced)&   N23&365K&0.123
& 0.080'& 0.524
&   7.47&1.23&  13.6
&2.35&\textbf{0.151}&\textbf{3.10}& 3.11& \textbf{0.282}&4.52
\\
Random forest&   WP22&47.5K&0.108
& 0.058'& 0.4'
&   9.20&2.65&  22.5
&2.52&0.502&5.04
 & 2.68& 0.543&5.40
\\
Cli-ViT&   N23&unavail.&0.086'
& 0.044'& 0.305
&   7.00&1.76&  15.8
&2.22&0.241&3.43
 & 2.80& 0.329&4.45
\\
ClimaX&   N23&108M&0.085'
& 0.043'& 0.297
&   \textbf{6.69}&\textbf{0.810}&  \textbf{10.7}&2.19&0.183&3.11
 & 2.68& 0.342&4.39
\\
Linear pattern scaling&   ours&27.7K&\textbf{0.0786}& \textbf{0.0410}& \textbf{0.284}&   8.02&2.15&  18.8&\textbf{1.87}&0.268&3.20& \textbf{2.25}& 0.357&\textbf{4.03}\\
\midrule
Target std. dev.&   WP22&-&0.052'&
 0.072'& 0.414&   2.51&1.49&  9.97&1.35&0.268& 2.69& 1.76& 0.457&4.04\\
\bottomrule
\end{tabular}}
\end{table}

\subsection{Analysing the relationship between internal variability and performance assessments of climate emulators}\label{sec:internal_var}
In this section, we analyse if internal variability contributes to the comparatively good scores of LPS on the ClimateBench spatial precipitation. We cannot conduct our internal variability experiment on the ClimateBench targets, because the experiment requires many ensemble members. Thus, we first demonstrate the magnitude of internal variability in the ClimateBench targets in comparison to the Em-MPI targets. Then, we show that the magnitude of internal variability in the targets affects the benchmark scores using the Em-MPI data \blue{and explain this influence using a simplified dataset}. This suggests that internal variability can impact benchmark scores on any dataset that does not contain enough members to average out internal variability, including the ClimateBench dataset, as discussed in~\cref{sec:discussion_int_var}.




\subsubsection{Magnitude of internal variability in 3-member NorESM2-LM and 50-member MPI-ESM1.2-LR ensemble average}\label{sec:em_mpi_internal_var}
\Cref{fig:internal_variability_graphic} illustrates the magnitude of internal variability in the 3-member NorESM2-LM targets in comparison to the 50-member MPI-ESM1.2-LR targets. 
First,~\cref{fig:internal_variability_tas} shows the ensemble-mean surface temperature anomalies over time for the NorESM (left) and Em-MPI (middle) data. The anomalies are plotted as a global-average (black) and regional-average over a sample region (red). 
Interannual fluctuations from internal variability are still visible in the global NorESM (left,black) data and are higher at the regional scale, exemplified by fluctuations up to $\pm 1.5 ^\circ C$ over the Greenland/Iceland region (left,red). Similar plots in~\cref{fig:internal_variability_pr} illustrate that internal variability is even higher for precipitation and dominates the signal on a regional scale, here over S. Central-America. 

The 50-member mean in the Em-MPI data (middle) shows notably lower fluctuations from internal variability for temperature and precipitation.~\Cref{fig:internal_variability_selected_regions} shows that the fluctuations from internal variability also seem lower across a diverse set of regions. 
Additionally, we plot averages over a 21-year sliding window in~\cref{fig:internal_variability_selected_regions}, because the spatial RMSE in~\cref{eq:rmse_spatial} computes a 21-year average before calculating the spatial errors. In most regions, the fluctuations in the 21-year averaged precipitation are also visibly lower in comparison to the NorESM2-LM ensemble-mean. 

\subsubsection{Effect of internal variability on benchmark scores}\label{sec:results_internal_variability}

\begin{figure}[t]
  \centering     
  \subfloat{
    \includegraphics[width=0.98\linewidth, trim={0 0 0 1.8cm}, clip]{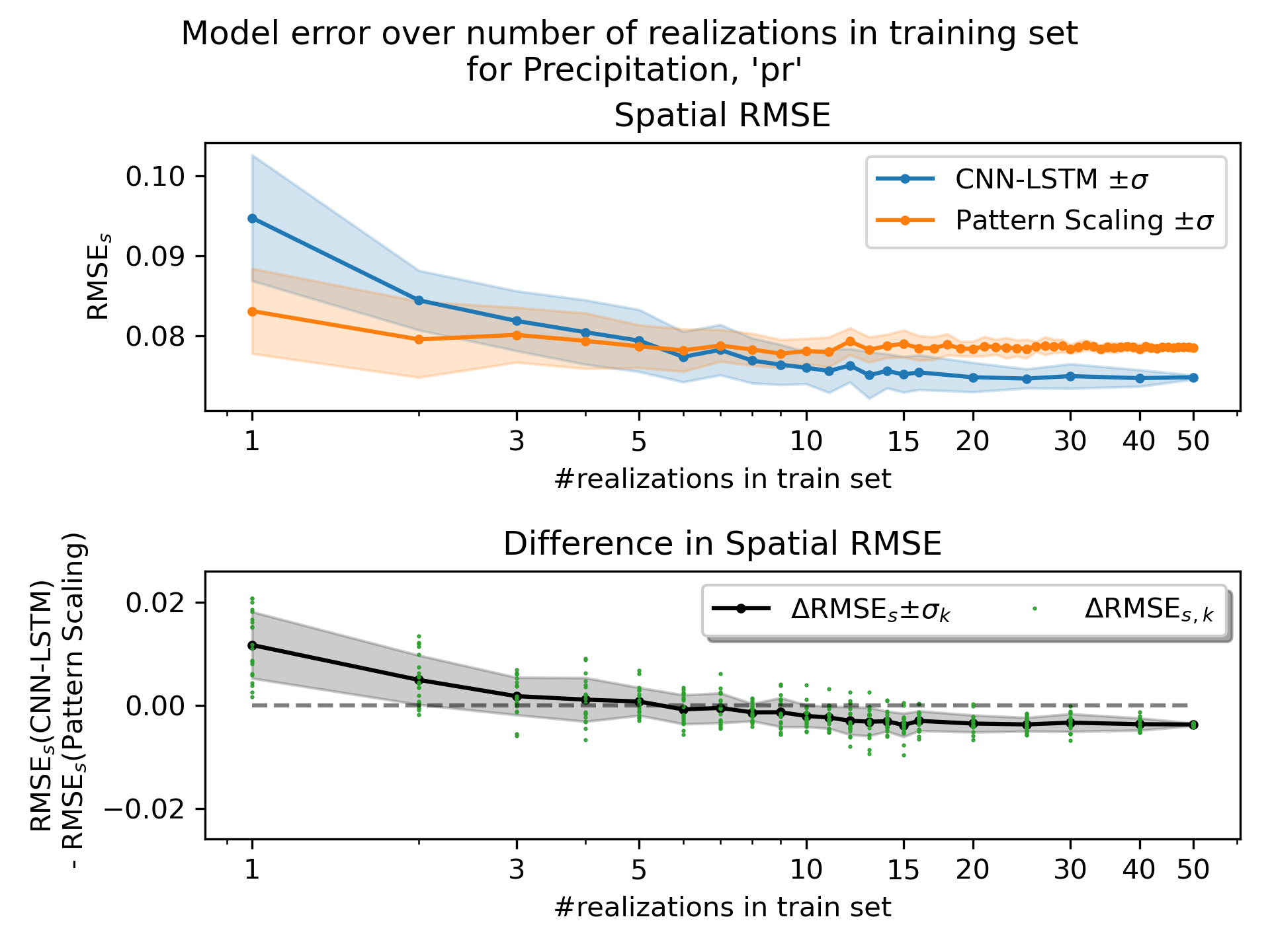}
  }
\caption[Precipitation over number of realizations]{\textbf{Error over realizations in training set for precipitation \blue{in mm/day}}. The top figure in our internal variability experiment shows the spatial RMSE, i.e., $\mathrm{RMSE}_{s,n}$, of an LPS (orange) and CNN-LSTM emulator (blue) that were trained on the ensemble-mean of data subsets with $n$ realizations and evaluated on the ensemble-mean of $N=50$ realizations from the Em-MPI data. Shading indicates the standard deviation across $K=20$ random draws of realization subsets. The bottom figure shows the difference in spatial RMSE, i.e., $\Delta \mathrm{RMSE}_{s,n}$, between the two emulators for each random realization subset, $(n,k)$, (green dots); and the mean and standard deviation across $K$ subsets (black line and shading). Data is plotted on a log x-axis, because most climate models are run for $1$ to $10$ realizations per scenario. As a side note, the $\pm \sigma$ range in the bottom figure is not the average of the two $\sigma$ ranges in the top figure, because an emulator's RMSE covaries with the subset, i.e., if one emulator has low RMSE the other emulator is also likely to have low RMSE. }
\label{fig:spatial_pr_over_realizations}
\end{figure}

~\Cref{fig:spatial_pr_over_realizations}-top shows the results of our internal variability experiment for precipitation anomalies from the Em-MPI data. When high internal variability is present in the training set at $n=3$ realizations, the LPS (orange) has a better spatial RMSE than the CNN-LSTM (blue). This mirrors the results on the ClimateBench results table in~\cref{tab:climatebench_pattern_scaling}. As the internal variability is increasingly averaged out by including more realizations in the training data set, 
the RMSE of the CNN-LSTM decreases. The LPS and CNN-LSTM RMSEs become comparable at $n=6$ realizations. For more realizations up to $50$ the CNN-LSTM has a better score than the LPS.
This trend is also visible when plotting the difference, $\Delta \text{RMSE}_s(n)$, which is positive at $n=3$ and crosses the zero-intercept at $n\approx6$ in~\cref{fig:spatial_pr_over_realizations}-bottom.

We illustrate that high internal variability skews the benchmarking scores towards a lower complexity emulator by fitting a linear trend to $\Delta \text{RMSE}(n)$ and observing a negative slope. The linear fits are shown in~\cref{fig:spatial_rmse_inset_over_realizations} and have a slope of $-5.77\times 10^{4}$ for spatial precipitation and a slope of $-12.2\times10^{4}$ for spatial surface temperature. 
We fit the linear trend only over the x-axis inset $n\in[0,20]$, because the subsets contain an increasing number of duplicates across $n$, which increasingly flattens out the slope even if there is still leftover internal variability, a known issue of ensemble analyses~\citep{tebaldi21extreme}. 
\strikeolive{We discuss possible reasons for internal variability affecting benchmarking scores in~\cref{sec:overfitting} and implications in~\cref{sec:discussion_int_var}.}

\Cref{fig:spatial_pr_over_realizations}-bottom also shows that there is at least one realization subset with $3$ realizations that has $\Delta \text{RMSE}_s<0$ (green dot below zero line). This implies that on some 3-member subsets from the MPI-ESM1.2-LR model, CNN-LSTM outperforms LPS on spatial precipitation.
And, this suggests that a lucky random draw of three realizations for the ClimateBench NorESM2-LM spatial precipitation targets could have also lead to a flipped result where the CNN-LSTM outperforms LPS.

\begin{figure}[t]
  \centering     
  \subfloat{
     \includegraphics[width=0.98\linewidth, trim={0 0 0 1.7cm}, clip]{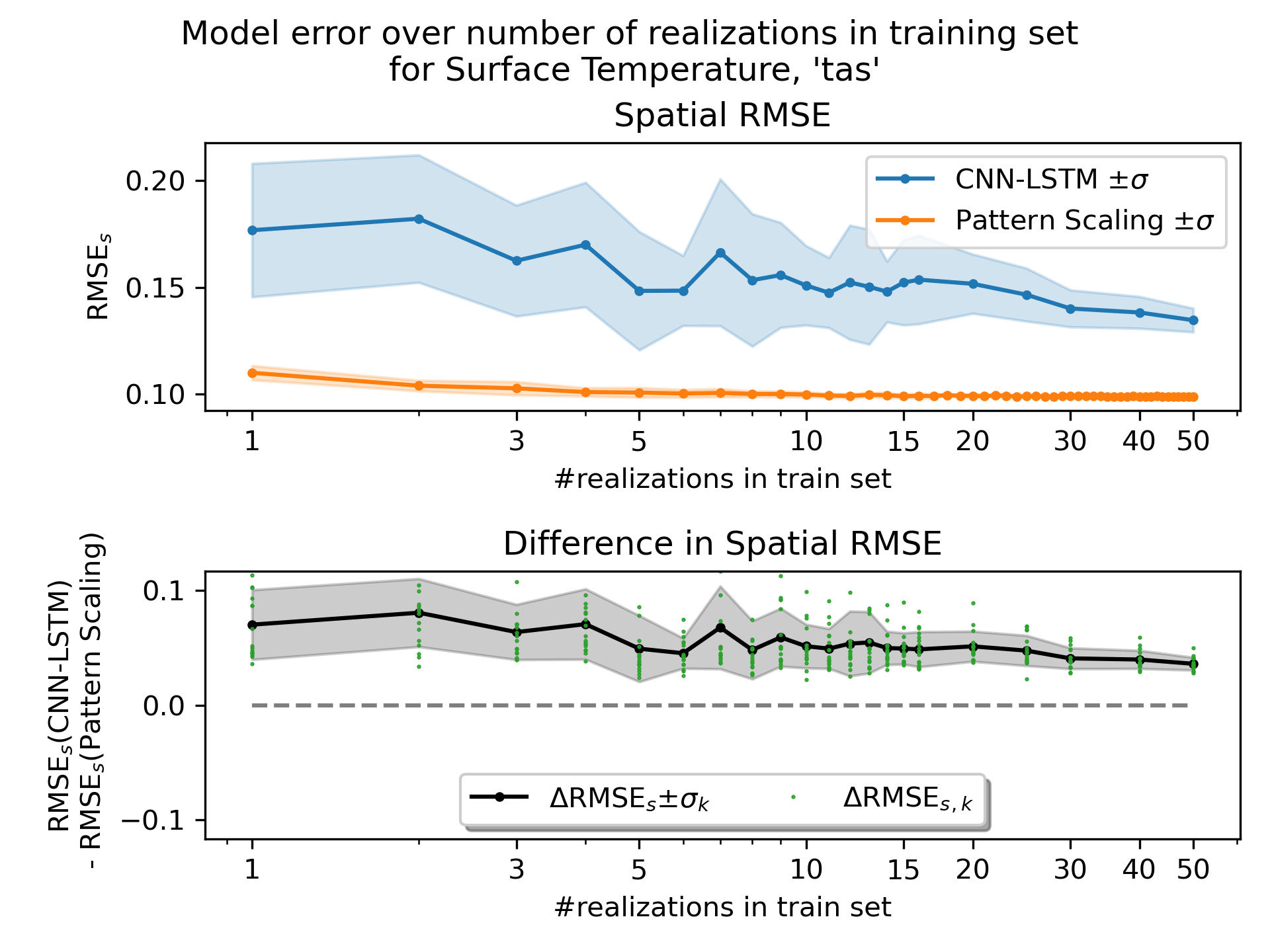}
  }
\caption[Spatial temperature over number of realizations]{\textbf{Error over realizations in training set for surface temperature \blue{\textdegree C}}. The LPS emulator (top-plot, orange) has a lower spatial RMSE on surface temperature in comparison to the CNN-LSTM (blue) independent of the number of realizations in the training set. The difference between both emulation techniques slightly decreases, but the scores between different subsets (bottom-plot, green dots) have significant spread.}
\label{fig:spatial_tas_over_realizations}
\end{figure}

\begin{table}
    \caption[Results table on Em-MPI target data]{\textbf{Em-MPI results table.} Spatial and global RMSE, following~\cref{eq:rmse_spatial,eq:rmse_global}, using the Em-MPI 50-member ensemble-mean as targets and the ClimateBench evaluation protocol. The scores contain three significant digits.}
    \label{tab:em_mpi_scoreboard}
    \centering
    \begin{tabular}{l lllcll}
    \toprule
         &  \#param&  \multicolumn{2}{c}{Surface temperature}& \multicolumn{2}{c}{Precipitation}\\
         \cmidrule(r){2-2}\cmidrule(r){3-4}\cmidrule(r){5-6}
         &  &  Spatial&Global& Spatial& Global\\
     \midrule
 CNN-LSTM               &$485$K & 0.135 &0.0679&\textbf{0.0748}&0.0152\\
 Linear pattern scaling &36.8K    & \textbf{0.0989}&\textbf{0.0265}&0.0785&0.0152\\
    \end{tabular}
\end{table}

For surface temperature, plotted in~\cref{fig:spatial_tas_over_realizations}, the LPS emulator achieves better spatial RMSE scores than the CNN-LSTM independent of the number of realizations in the training set. The RMSE scores at $n=50$ are given in~\cref{tab:em_mpi_scoreboard}. Similar to precipitation, the difference between LPS and CNN-LSTM (shown in~\cref{fig:spatial_tas_over_realizations}-bottom) decreases with increasing number of realizations in the training set although with a negative slope of lower magnitude. These results would be in line with an expectation that surface temperature contains less internal variability than precipitation and is thus averaged out with fewer climate realizations~\citep{tebaldi21extreme}. Further, these results are consistent with our expectation given the predominantly linear relationship of surface temperature with cumulative $\mathrm{CO}_2$ emissions.

\subsubsection{Explanation of overfitting and the bias-variance tradeoff}\label{sec:results_overfitting}

\begin{figure}[t]
  \centering     
  \subfloat[Mean of 2 realizations]{
    \includegraphics[width=0.3\linewidth]{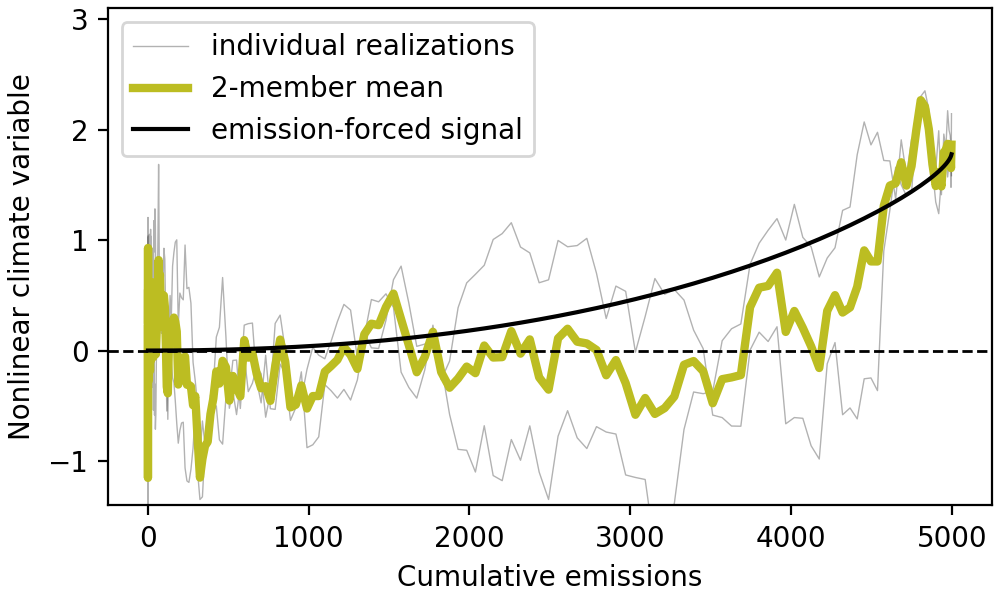}
    \label{fig:nonlinear_climate_over_cumul_emissions}
  }
  \subfloat[Fits on 2-member train set]{
    \includegraphics[width=0.3\linewidth]{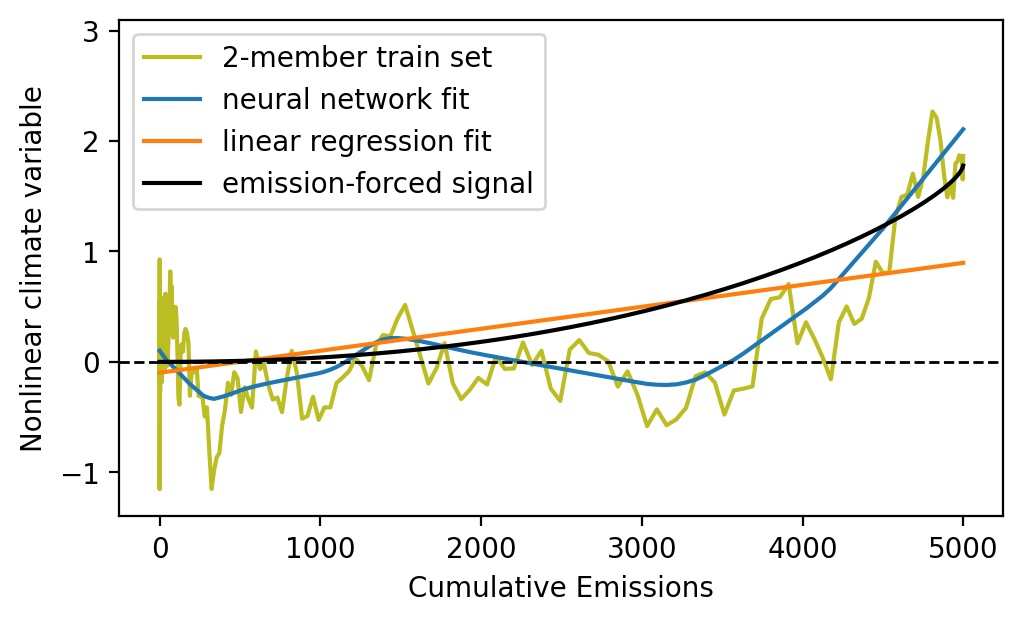}
    \label{fig:single_draw_w_2members}
  }
  \subfloat[Fourier spectra]{
    \includegraphics[width=0.33\linewidth, trim={0 0.4cm 0 0.4cm}, clip]{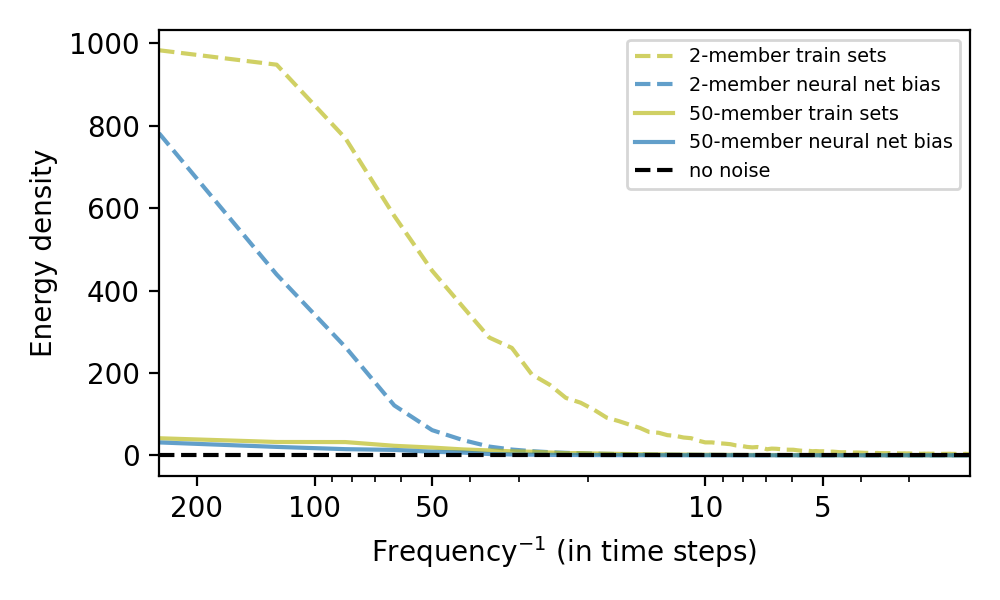}
    \label{fig:fourier}
  }\\

  \subfloat[Linear regression fits]{
    \includegraphics[width=0.3\linewidth]{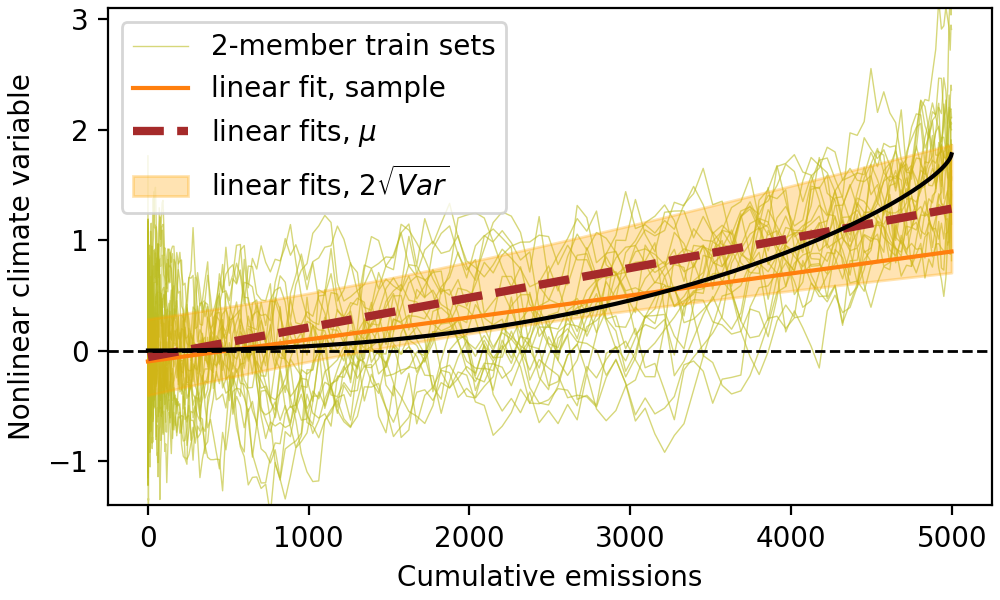}
    \label{fig:linear_fits_cumul_emissions_2members}
  }
  \subfloat[Neural network fits]{
    \includegraphics[width=0.3\linewidth]{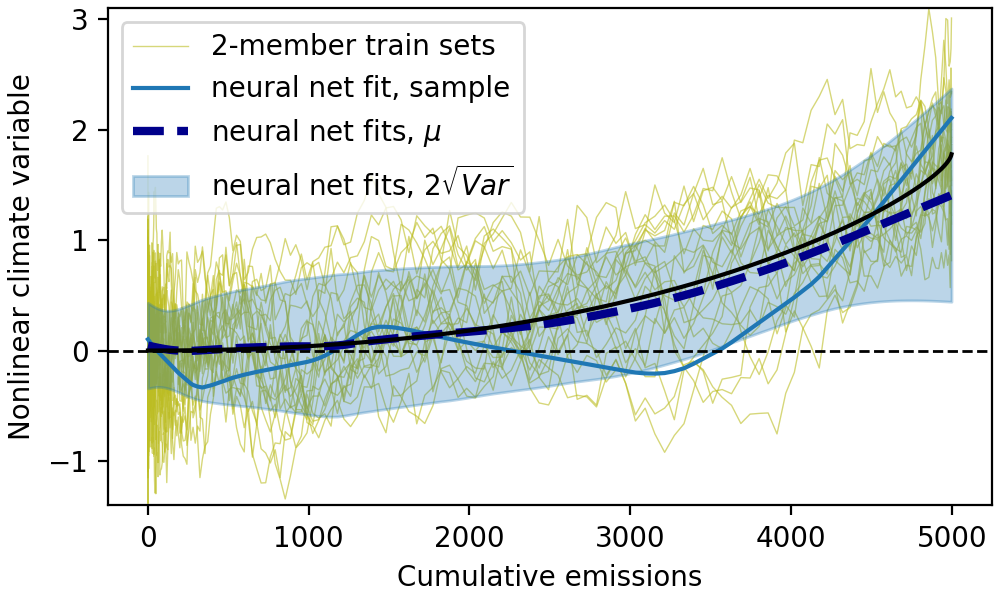}
    \label{fig:fcn_fits_cumul_emissions_2members}
  }
  \subfloat[MSE, bias, and variance]{
    \includegraphics[width=0.31\linewidth]{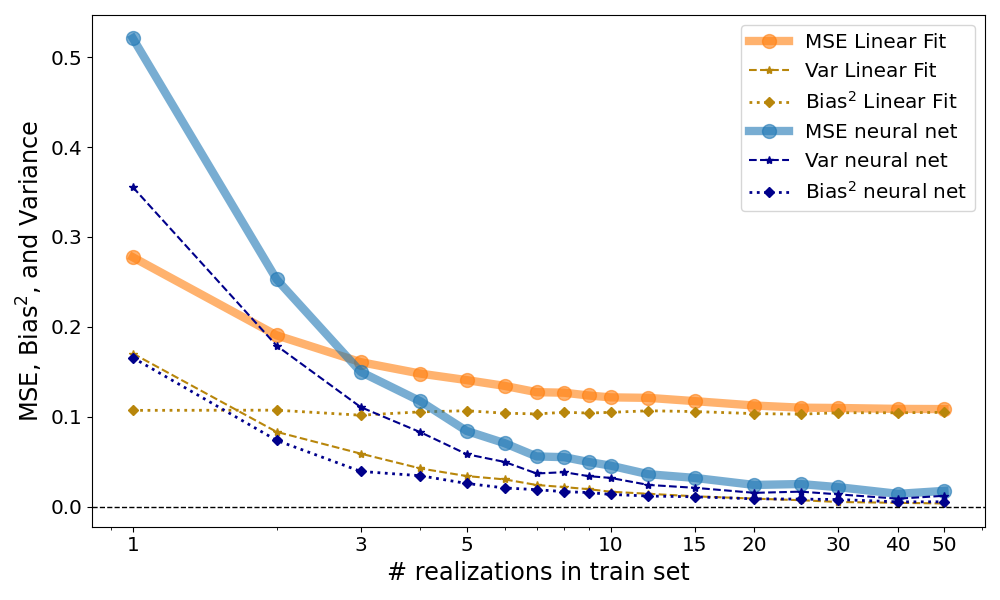}
    \label{fig:bias_var_ou_decomposition}
  }
\caption[Overfitting experiment]{\blue{\textbf{Bias-variance trade-off on a simplified dataset}.~\Cref{fig:nonlinear_climate_over_cumul_emissions} shows two realizations (gray lines), the corresponding ensemble-mean (olive), and the emission-forced signal (black) of an idealized one-dimensional stochastic function.
The best neural network (blue) and linear (orange) fit to this $2$-member ensemble-mean are plotted in~\Cref{fig:single_draw_w_2members}.~\Cref{fig:linear_fits_cumul_emissions_2members,fig:fcn_fits_cumul_emissions_2members} show the mean and variance of each emulation technique across $500$x draws of a $2$-member ensemble-mean, illustrating a comparatively higher variance across neural network fits.~\Cref{fig:fourier} shows the expected Fourier spectra of an $n$-member average of the underlying noise process and the corresponding signal-removed neural network fits, according to~\cref{eq:fourier_spectra}.~\Cref{fig:bias_var_ou_decomposition} plots the expected variance (star-dashed), bias$^2$ (diamond-dashed), and MSE (shaded line) of each emulation techniques as a function of members in the ensemble-mean training sets.
}}
\label{fig:ou_targets}
\end{figure}

\blue{
The comparatively worse performance of a deep learning emulator in the noisy data regime (low $n$) can be explained on the simplified dataset.~\Cref{fig:single_draw_w_2members} illustrates that a neural network fit (blue) can be influenced by the low-frequency variability in a training set that only contains $2$ members (olive).
Plotting the Fourier spectra of neural net biases (blue) and the noise-process (olive) in~\cref{fig:fourier} shows that the neural network fits on $2$-member training sets maintain some energy from the noise process at lower frequencies, while filtering out higher frequency noise. If a neural network would have accurately filtered out all noise, the energy density would be close to zero in this plot (see $50$-member fits). 
This shows that emulation techniques can overfit to noise in the training data, noting that the range of overfitted frequencies should vary depending on the number of free parameters and regularization techniques. 
}

\blue{
The bias-variance tradeoff gives an explanation why the highest-scoring emulation technique switches as a function of $n$: When there is significant noise in the training set there is more variance between different neural network fits than linear fits, as illustrated with $2$-member sets in~\cref{fig:linear_fits_cumul_emissions_2members,fig:fcn_fits_cumul_emissions_2members}. Plotting the variance over $n$ in~\cref{fig:bias_var_ou_decomposition} confirms this quantitatively and shows that the variance (purple, star-dashed) is the major contributor to the neural networks' MSE score (blue, solid) at $n{=}2$.
The linear fits have lower variance at $n{=}2$ (brown, star-dashed), but also take a strong assumption on the functional form and, thus, have a significant bias (diamond-dashed, brown). 
As the number of realizations in the training set increases, the bias of the linear fits stays constantly high whereas the neural networks' variance decreases. This causes a switch of the emulator with the lower MSE score at $\sim 3$ realizations, similar to the switch of the higher scoring emulator on the Em-MPI spatial precipitation targets. 

The bias-variance tradeoff also explains the effect of regularization. In this case, the neural network is regularized causing a non-zero bias (diamond-dashed, brown). But, retraining the neural network without regularization would result in a zero bias and higher variance, e.g., at $n=2$~\citep{bishop06bishop}.
}

\section{Discussion}\label{sec:discussion}


\subsection{Importance of comparing with an LPS baseline}
\label{sec:lps_linearity_discussion}\label{sec:lps_limitations}

Our analysis indicates that quantitative comparisons with LPS can help determine if deep learning-based emulators are making meaningful progress on known issues in climate emulation. For example, it is known that the relationships between global annual mean surface temperature vs. cumulative $\mathrm{CO}_2$ emissions~\citep{macdougall16tcre} and local vs. global annual mean surface temperature~\citep{tebaldi14patternscalingstrengths} are predominantly linear and these relationships could be seen as minimum criteria for validating deep learning-based emulators. 
Deep-learning-based emulators may hold more promise for emulating settings that are known limitations of 
LPS, such as emulating overshoot scenarios, daily statistics \blue{and hourly dynamics}, abrupt transitions, or extremes in regions with narrow temperature distributions~\citep{womack25responsefunction,macdougall16tcre,wells23patternscaling,tebaldi20patternscalingextreme}. Further, nonlocal aerosols~\citep{levy13aerosols,williams22nonlocalaerosols,williams23nonlinearaerosolsprecip}, \blue{short-term effects of $\mathrm{CO}_2$~\citep{cao15fastco2response}}, or convection over tropical oceans~\citep{kravitz17precippatternscaling} can impact precipitation nonlinearly. \blue{Theoretical insights on the limitations of linear emulators can be found in literature surrounding the fluctuation-dissipation theorem~\citep{leith75fdt,lembo20fdtemulator} and an analysis by~\citet{giani24patternscaling}}. Our work has shown that a deep learning-based emulator can improve marginally on LPS for emulating the Em-MPI spatial precipitation targets (see \cref{tab:em_mpi_scoreboard}). However, many nonlinear phenomena remain to be studied to determine if advanced emulation techniques could improve on LPS baselines.

Generally speaking, our study emphasizes the importance of benchmarking machine learning (ML)-based emulators against established techniques. While the allure of novel ML approaches is understandable, 
it is not a good reason to neglect well-developed methods such as LPS. Neglecting well-developed methods may lead to an overestimate ML's skill, such as in the case of transformer-based emulators being cited as the most accurate emulators of surface temperature.
Our work serves as a cautionary tale, highlighting the possibility of overlooking simpler techniques that may achieve comparable or superior performance. 

\subsection{Overfitting to internal variability and the bias-variance trade-off}\label{sec:overfitting}
\blue{Our results have shown that the highest-scoring emulation technique (LPS vs. CNN-LSTM) can change depending on the number of realizations (see~\cref{fig:spatial_pr_over_realizations}). This is perhaps surprising if the underlying relationship is nonlinear in which case a CNN-LSTM would be expected to outperform a linear model~\citep{bishop06bishop}. But, using the simplified dataset in~\cref{fig:ou_targets}, we have shown that a deep learning-based emulator can also have comparatively high variance, leading to the emulator overfitting to low-frequency variability. And, if this variance outweighs the bias that is introduced through assuming a linear relationship, then linear regression can be the better scoring emulation technique to nonlinear functions.}

\blue{This indicates that a general-purpose technique for emulating any variable of any ESM would need to be flexible in trading off bias and variance. In particular, this becomes relevant in the low-data regime, which can be the case for ESMs that only contain a few realizations per scenario and climate variables, such as the change in surface winds, that have a low signal-to-noise ratio. Neural networks are a universal function approximator and, thus, appealing as general-purpose technique. But, at least in the case of our simplified dataset, standard methods of using weight decay and early stopping were insufficient to prevent neural networks from overfitting to low-frequency variability in this low-data regime. Thus, we recommend to additionally regularize emulation techniques through climate science-driven assumptions on the functional form and emulator's complexity. Testing out various assumptions can be part of a hyperparameter optimization process that is conducted per-variable, per-ESM, recognizing that the highest scoring complexity may be quite simple. 

\blue{It is an open question which choice of validation set would be the most practical for identifying this optimal complexity in future climate emulation benchmarks. In our internal variability experiment, we did not withhold a validation set for consistency with ClimateBench. 
For a future climate emulation benchmark}
with a small number of emission scenarios (e.g., 5) an alternative option would be to use a strategy of cross-validation similar to~\citep{bouabid24fairgp}: 
train on 3 and validate on 1 scenario, optimize the hyperparameters on all 3:1 combinations of scenarios, test the model on 1 completely held out scenario, and repeat the steps for all held out scenarios that may be of scientific interest. Nevertheless, when only a few realizations per scenario are available it continues to depend on each climate variable's signal-to-noise ratio if the set of all realizations across all scenarios contains sufficient information for distinguishing random low-frequency fluctuations, such as multidecadal oscillations, from long-term emission-forced signals, independent of how the validation set is chosen. 
}
\strikeolive{
Three factors -- model complexity, the training objective, and the data split -- contributed to the CNN-LSTM failure and could be adjusted for future benchmarking. First, the results in~\cref{fig:spatial_pr_over_realizations} suggest that the model with high model complexity (CNN-LSTM), i.e., many parameters, may overfit on the noise in the training targets at low $n$. In the internal variability experiment at low $n$, emulators are trained on ensemble-averages that still contain ``noise'' from strong interannual fluctuations, but evaluated on data with smooth trajectories. 
Given noisy data ($n=3$), emulators with high model complexity are likely to overfit without regularization techniques, such as weight decay~\citep{bishop06bishop}. In comparison, emulators with low model complexity, such as LPS, give predictions that are likely underfitted but filter out noise by design~\citep{bishop06bishop}. 
In the case of spatial precipitation once the noise is decreased ($n>=6$), the CNN-LSTM can achieve higher accuracy on interpolation tasks, likely because the signal becomes smoother and overfitting becomes less of a limitation than LPS' inability to emulate the nonlinear behavior of precipitation.
Future comparisons with deep learning models could address the issue of overfitting with regularization techniques or choosing model architectures with less parameters~\citep{goodfellow16deeplearning}.
}

\strikeolive{
Second, the training objective in the deep learning-based CNN-LSTM and ClimaX emulator is to minimize a per-year MSE (detailed for the CNN-LSTM in~\cref{eq:cnn_lstm_loss}), which differs from the spatial 21-year average evaluation objective in~\cref{eq:rmse_spatial}. The per-year MSE encourages the emulator to learn interannual fluctuations that are more noisy than in a 21-year average. Smoothing the training targets over 21-year windows before training would reduce the noise which could help in lowering the risk of overfitting on interannual fluctuations. Further research could investigate if smoothing targets before training would lead to better performance of deep learning emulators when only a few climate realizations are available.
}

\strikeolive{Third, the evaluation protocol we follow splits the data into an identical training and validation set. Overfitting on small datasets can be limited by detecting it and using early stopping, regularization, or a lower model complexity~\citep{goodfellow16deeplearning}. 
Scientists usually monitor if the loss curves between training and validation sets are diverging to detect overfitting, but this becomes unfeasible when these data splits are not distinct~\citep{goodfellow16deeplearning}. 
In future benchmarks with a small number of emission scenarios (e.g., 5), a strategy of cross-validation similar to~\citet{bouabid24fairgp} might help: train on 3 and validate on 1 scenario, optimize the hyperparameters on all 3:1 combinations of scenarios, test the model on 1 completely held out scenario, and repeat the steps for all held out scenarios that may be of scientific interest.}


\subsection{Limitations and applicability of our results}\label{sec:discussion_int_var}

The difference in spatial precipitation RMSE between the two emulation techniques is relatively low ($\lvert \Delta \text{RMSE}_s\rvert < 0.015 \text{mm/day} \;\forall n\in\{1,...,N\}$). In this context, we note that justifications to switch from LPS to a deep learning-based emulator will depend on the application and likely require additional application-oriented evaluation metrics.
But, independent of the error magnitude, our work shows that reporting the difference on an average of $50$ instead of $3$ realizations will give a more robust estimate of an emulator's accuracy. 

While we illustrated the impact of internal variability only on emulators of spatial patterns of temperature and precipitation in the MPI-ESM1.2-LR model, we expect similar results to apply to many other variables and models.
In particular, we would expect that the benchmark scores on the NorESM2-LM 3-member mean spatial precipitation targets are also affected by internal variability due to the illustrated high magnitude of internal variability. 
Further, regionally-resolved near-surface wind velocities, for example, are known to have higher internal variability and a lower signal-to-noise ratio than precipitation which would raise the risk of overfitting. In general, if an emulation technique with high model complexity is overfitting when trained on the average of a  few $n$ realizations (which is not given), we would expect a downward trend in $\Delta \text{RMSE}_s(n)$ with increasing $n$ (as in~\cref{fig:spatial_pr_over_realizations}-bottom and~\cref{fig:spatial_tas_over_realizations}-bottom), although, the curvature and zero-intercept (if any) would change as a function of internal variability, linearity, and signal-to-noise ratio in the climate variable of interest.

\blue{
To provide further empirical evidence for this relationship, LPS could also be compared with the deep learning-based emulation techniques in ClimateSet~\citep{kaltenborn23climateset}. ClimateSet is an alternative dataset for evaluating emulation techniques that is similar to ClimateBench, but uses monthly-averaged climate variables (tas, pr) and projections from multiple climate models as targets. As the test protocol in ClimateSet only uses a single realization per climate model, many of our results on ClimateBench likely also apply to ClimateSet. 
Unfortunately, the results in ClimateSet were not reproducible at the time of writing our paper~\citep{lutjens24climatesetgithub}, so we refer to future work for this comparison.}



There are other ways of separating the forced response from internal variability beyond using more realizations and it would be interesting to study their 
relative advantages for assessing benchmarks.
This becomes especially relevant for emulating high-resolution climate data ($<100km$) and scenarios for which models only contain a few realizations.
For example, one can train or evaluate the emulator on variables averaged over a decadal time window. However, even a 30-year average can be insufficient to average out internal variability on a single climate realization~\citep{maher20internalvariability}. 
Alternatively, approaches from statistical learning~\citep{sippel19statisticallearningvariability} or pattern recognition~\citep{wills20patternrecognitionvariability} 
are other promising avenues for extracting the forced target signal from emission scenarios with only few realizations.

\section{Outlook and conclusion}\label{sec:future_benchmarks}


Our work aims to improve the evaluation process of climate emulators by focusing on the impacts of internal variability. Besides internal variability, however, there are interesting directions towards improving climate emulation benchmarking (e.g., for ClimateBenchv2.0): 
\blue{First, additional scenarios, such as single forcer (4xCO2) or out-of-distribution (ssp119) experiments, could cover a less correlated input space and be used to evaluate an emulator's capability to extrapolate beyond the training data.} 
Second, the current evaluation metric, NRMSE, can be dominated by local extremes and only measures the accuracy of \strikeolive{a} the predicted ensemble-mean. A future benchmark could add evaluation metrics that are more application-oriented (e.g., drought indices) or focus on assessing stochastic techniques, such as weather generators or diffusion models, that may aim to emulate the distribution of future climate states. 
A new benchmark could also include more diverse climate variables (e.g., highly varying regional winds or changes beyond 2100), a comparison with existing  emulation techniques beyond LPS (e.g.,~\citep{beusch22mesmermagicc} or ~\citep{dorheim23hectorv3}), and a cross-validation strategy.

In conclusion, we showed that linear pattern scaling emulates the local annual mean temperature, precipitation, and extreme precipitation in ClimateBench more accurately than current deep learning emulators. We identified internal variability as a key reason for the comparatively good performance of linear pattern scaling on precipitation. This implies that addressing internal variability is necessary for benchmarking climate emulators.
We assembled a data subset, Em-MPI, from the MPI-ESM1.2-LR ensemble that contains 50 instead of 3 realizations and can be used to augment the NorESM2-LM temperature and precipitation datasets in ClimateBench. Using the new dataset we showed that a CNN-LSTM can be more accurate for emulating precipitation, while linear pattern scaling continues to be the more accurate emulator for surface temperature anomalies, in terms of spatial RMSE scores.

\section*{Reproducibility and data availability statement}
The Em-MPI dataset is published with a CC-BY-4.0 license at \href{https://huggingface.co/datasets/blutjens/em-mpi}{huggingface.co/ datasets/blutjens/em-mpi}. The code for reproducing all results is published at \href{https://github.com/blutjens/climate-emulator}{github.com/ blutjens/climate-emulator}. The MPI-ESM1.2-LR data was retrieved from the Earth System Grid Federation interface~\citep{hermann19mpiesgf} under a CC-BY-4.0 license via the ESGF Pyclient API and building off the code in the ClimateSet downloader~\citep{kaltenborn23climateset}. The ClimateBench dataset was retrieved from~\citep{watsonparris22climatebenchzenodo} under a CC BY 4.0 license using the Python code in our github repository.

The internal variability experiment was implemented on a CPU cluster instead of GPUs, due to the parallelizability of the experiment and limited GPU availability; the experiment took a total of two weeks on 400 CPU cores. We used a mix of Intel Xeon Gold 6336Y (2.4GHz) and E5-2670 (2.6GHz). 





\authorcontributions
B.L. is the lead author and contributed to all CRediT author roles except for funding acquisition. D. W.-P. contributed to conceptualization and writing - review and editing. R.F. and N.S. contributed to conceptualization, methodology, validation, resources, writing - review and editing, and funding acquisition. 

\acknowledgments
This project is supported by Schmidt Sciences, LLC and part of the MIT Climate Grand Challenges team Bringing Computation to the Climate Challenge (BC3).


Claudia Tebaldi has been instrumental for her insight on statistical analyses and existing literature in emulation. We are very grateful to Andre N. Souza, Paolo Giani and the BC3 team members for regular methodological feedback and discussions regarding this work. We thank Konstantin Klemmer, Ruizhe Huang, Mengze Wang, and Chris Womack for their thorough pre-submission reviews and appreciate the dialogues with Iris de Vries and Shahine Bouabid. Thank you, to Dava Newman for the inspirational support to start this line of work. Thank you, also to Katharina Berger for the rapid responses to support downloading the MPI data. \blue{We deeply appreciate the extensive feedback that we have received during the anonymous peer-review process.}
\bibliography{references}

\clearpage
\appendix
\section{Definitions}\label{app:definitions}

\subsection{Background on internal variability}\label{sec:internal_var_for_ml}
Internal variability refers to natural fluctuations in weather and climate~\citep{leith78predictability}. \blue{These fluctuations are often decomposed into modes with periods of varying lengths, such as the Madden-Julian Oscillation (1-3 months), El Ni\~no Southern Oscillation (2-7 years), or Atlantic Multidecadal Oscillation (60-70 years).} The magnitude of the associated amplitudes can be at the same scale or higher than the emission-induced climate change, depending on the climate variable of interest. It is known that the exact transition times of some of those climate modes, such as ENSO, are unpredictable on climate timescales~\citep{tang18ensopredictability}. Thus, climate scientists often treat the fluctuations as "climate noise" and identify the forced signal by taking an ensemble average over a sufficiently large set of climate simulations~\citep{maher19mpige}. These ensembles can be called single model initial-condition ensembles (SMILEs) and contain multiple realizations, i.e., simulations, runs, or members, of a single model that is forced by the same emission pathway but with slightly different initial conditions on every run. 

\section{Extended Data \& Methods}\label{app:dataset}
\subsection{Appendix to Background on ClimateBench dataset, evaluation protocol, and metrics}\label{app:background_climatebench}
\subsubsection{Definition of variables in ClimateBench inputs and targets}\label{app:climatebench}
\begin{figure}[ht]
  \centering     
  \subfloat{
    \includegraphics[width=0.98\linewidth]{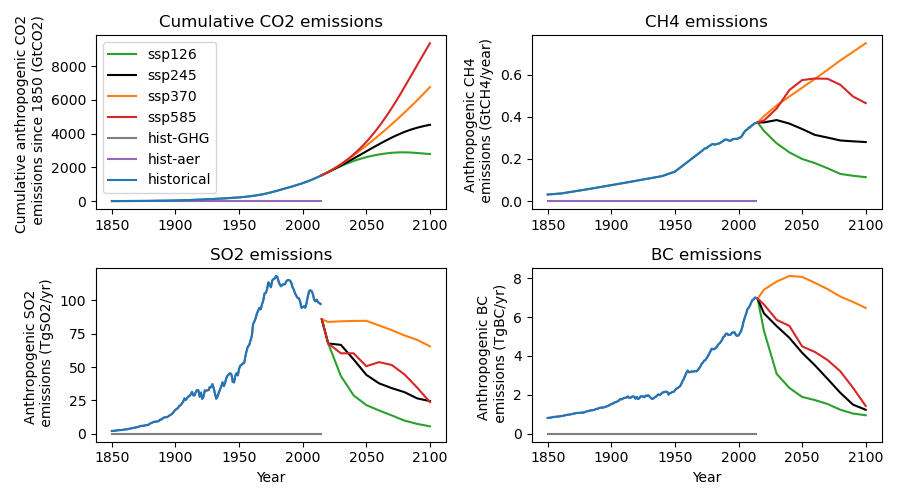}
    \vspace{-0.1in}
  }
\caption[]{\textbf{Emission pathways over time.} The plots show multiple emission pathways for each forcing variable. The $\mathrm{CO}_2$ emissions (top-left) are plotted as sums over the globe and years since 1850. The methane (top-right), sulfur-dioxide (bottom-left), and black carbon (bottom-right) are plotted as global sums per year. $\mathrm{SO}_2$ contains a discontinuity between historical and ssp scenarios. The ClimateBench inputs contain global values for $\mathrm{CO}_2^\text{cum}$ and $\mathrm{CH}_4$ and locally-resolved values for $\mathrm{SO}_2$ and $BC$ (not shown).}
\label{fig:emissions_over_time}
\end{figure}

\begin{figure}[t]
  \centering     
  \subfloat{
    \includegraphics[width=0.98\linewidth]{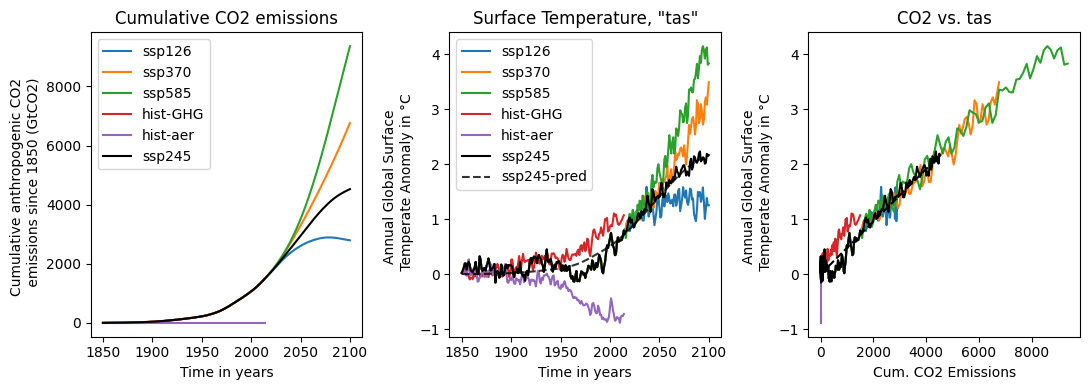}
  }
\caption[Temperature and CO2 over time]{\textbf{Correlation between temperature and global $\mathrm{CO}_2^{cum}$ with 3-member NorESM2-LM targets}. The left plot shows the input cumulative $\mathrm{CO}_2$ emissions over time for each training scenario (ssp126, -370, -585, hist-GHG, hist-aer) and test scenario (ssp245), averaged over three realizations of the NorESM2-LM model. The middle plot shows the annual globally-averaged surface temperature anomalies, wrt. the preindustrial control, over time for each scenario. The black-dotted line indicates the linear pattern scaling emulator prediction. On the right, we plot the data against each other. We can see that global temperature and $\mathrm{CO}_2^{cum}$ are linearly correlated except for the high-frequency variations, the hist-aer scenario that ablates $\mathrm{CO}_2^{cum}$, and a dip in 1970-2000.}
\label{fig:tas_co2_over_time}
\end{figure}

The variables are defined as:
\begin{itemize}
    \item $\mathrm{CO}_{2;t}$ are the carbon dioxide emissions, from anthropogenic sources, summed over the globe and the year, $t$. The cumulative carbon dioxide emissions, $\mathrm{CO}_{2;t}^{cum}$, are summed from 1850 until the end of the target year, $t$. We visualize the data in gigatons of carbon dioxide, GtCO2, (one gigaton is one billion tons) in~\cref{fig:emissions_over_time}.
    \item $\mathrm{CH}_{4,t}$ are the anthropogenic methane emissions summed over the globe and year, $t$.
    \item $\mathrm{SO}_{2;i,j,t}$ are the anthropogenic sulfur-dioxide emissions summed over the area of grid cell $(i,j)$ and year, $t$.~\Cref{fig:emissions_over_time} shows that the emission pathways for $\mathrm{SO}_2$ contain a discontinuity between historical and ssp scenarios at year 2014-15. This discontinuity exists in the official CMIP6 data and has been updated in the 2022 CEDS data, as illustrated in~\citep{smith21so2discontinuity}-Fig. S2, but the NorESM2-LM and Em-MPI targets were still generated with the official CMIP6 forcings. 
    \item $\mathrm{BC}_{i,j,t}$ are the anthropogenic black carbon emissions summed over the area of grid point $(i,j)$ and year, $t$. 
    \item $\mathrm{tas}_{i,j,t}$ are the annual mean surface temperature anomalies at $2\mathrm{m}$ height. The surface temperature anomaly is the surface temperature minus the average surface temperature of the preindustrial control simulation at each grid point. The unit is $\deg C$. The globally-averaged values are plotted in~\cref{fig:tas_co2_over_time}.
    \item $\mathrm{dtr}_{i,j,t}$ is the annual mean diurnal temperature range anomaly wrt. the preindustrial control run in $\deg C$. The diurnal temperature range was calculated by first subtracting the minimum from the maximum daily temperature and then averaging over the year.
    \item $\mathrm{pr}_{i,j,t}$ is the annual mean total precipitation per day in $mm/day$. The value is reported as anomaly wrt. to the preindustrial control run. 
    \item $\mathrm{pr90}_{i,j,t}$ is the 90th percentile of the daily precipitation in each year. 
    \item We refer to~\citet{watsonparris21climatebench} for more details.
\end{itemize}

To analyze the linearity we have plotted the correlation between temperature and global $\mathrm{CO}_2$ in~\cref{fig:tas_co2_over_time}. 

\subsubsection{Evaluation metrics in ClimateBench }\label{app:climatebench_metrics}
For the results table in ClimateBench (see~\cref{tab:climatebench_pattern_scaling}), the RMSE is normalized by the magnitude of the target variable to compute the NRMSEs, where we denote the placeholder $*\in\{s,g\}$:

\begin{equation}
    \mathrm{NRMSE}_*(\hat \tY_e, \tY_e) = \frac{\mathrm{RMSE}_*}{\left|\frac{1}{\sum_i \alpha_i J}\sum_{i,j}\alpha_i\frac{1}{\lvert\sT_\text{test}\rvert}\sum_{t} \frac{1}{\lvert\sM\rvert} \sum_m y_{i,j,t,m,e}\right|_\text{abs}}
\label{eq:nrmse_spatial}
\end{equation}

The ClimateBench evaluation protocol and results table also contain the weighted sum of $5\times$ global + spatial NRMSE, called "total" NRMSE. The total NRMSE was proposed in the ClimateBench paper as a metric on which the "best" emulator can be declared and the factor of $5$ was chosen to equalize the scales between both errors.

The ClimateBench evaluation metrics in~\cref{eq:rmse_spatial,eq:rmse_global,eq:nrmse_spatial} are consistent with the ClimateBench paper Sect. 3.1. Eq. (1-3) except for a slightly different notation.

\subsection{Appendix to Em-MPI data}\label{app:em_mpi}
We followed the~\citet{hausfather22hotmodel} recommendation to avoid using "hot" models and decided against using CanESM5 with the transient climate response, TCR${\approx} 2.7^\circ$, being outside the $[1.4,2.2]^\circ C$ likely range and the equilibrium climate sensitivity, ECS${\approx} 5.6^\circ C$, being outside the $[2.3,4.7]^\circ C$ very likely range~\citep{tokarska20tcrecs,sherwood20likelyecs}.
\blue{However, excluding hot models from climate analyses is an ongoing point of discussion, especially for regional climate analysis~\citep{swaminathan24hotmodels}, and CanESM5 might provide a valuable reference for future benchmarking of emulators.} We decided against using EC-Earth3 because many realizations were unavailable on the Earth System Grid Federation (ESGF) Portal\blue{, but they may be retrievable through other pathways}. For MPI-ESM1.2-LR all realizations were either already available or rapidly made available on ESGF after inquiry with the data owners. \blue{The MMLEAv2 was unavailable at the start of this work, but contains 30+ realizations of the Tier 1 SSPs for a few additional ESMs~\citep{maher24mmleav2}.}

\begin{figure}[ht]
  \centering     
  \subfloat[Spatial distributions of precipitation anomalies. Each histogram shows the distribution of annually- and ensemble-averaged precipitation anomalies over all grid points and years in the selected timeframe for a given scenario. We select short time frames  -- 1850-1900 for historical and 2080-2100 for the ssp-scenarios -- to increase the sample size but maintain minimal influence from changing emissions.]{
    \includegraphics[width=0.97\linewidth, trim={0 0 0 1cm}, clip]{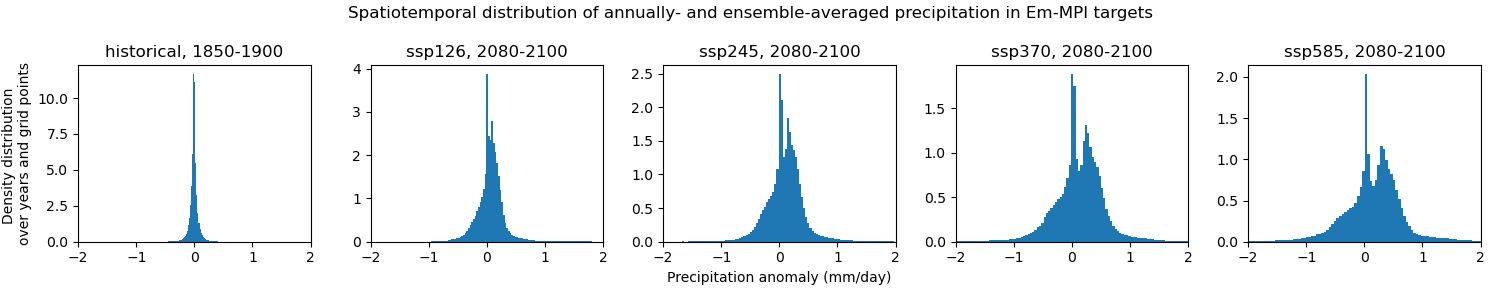}
    \label{fig:mpi_distribution_pr_over_loc_and_year}
  }\\
  \subfloat[Ensemble distributions of precipitation anomalies. 
  Each histogram shows the distribution of annually-averaged precipitation anomalies over all 50 members in the ensemble, and years in a subset of the historical period (1850-1900). \blue{Each column shows the distributions for a different region, and each row shows a different randomly selected grid cell within this region.}]{
    \includegraphics[width=0.97\linewidth, trim={0 0 0 0}, clip]{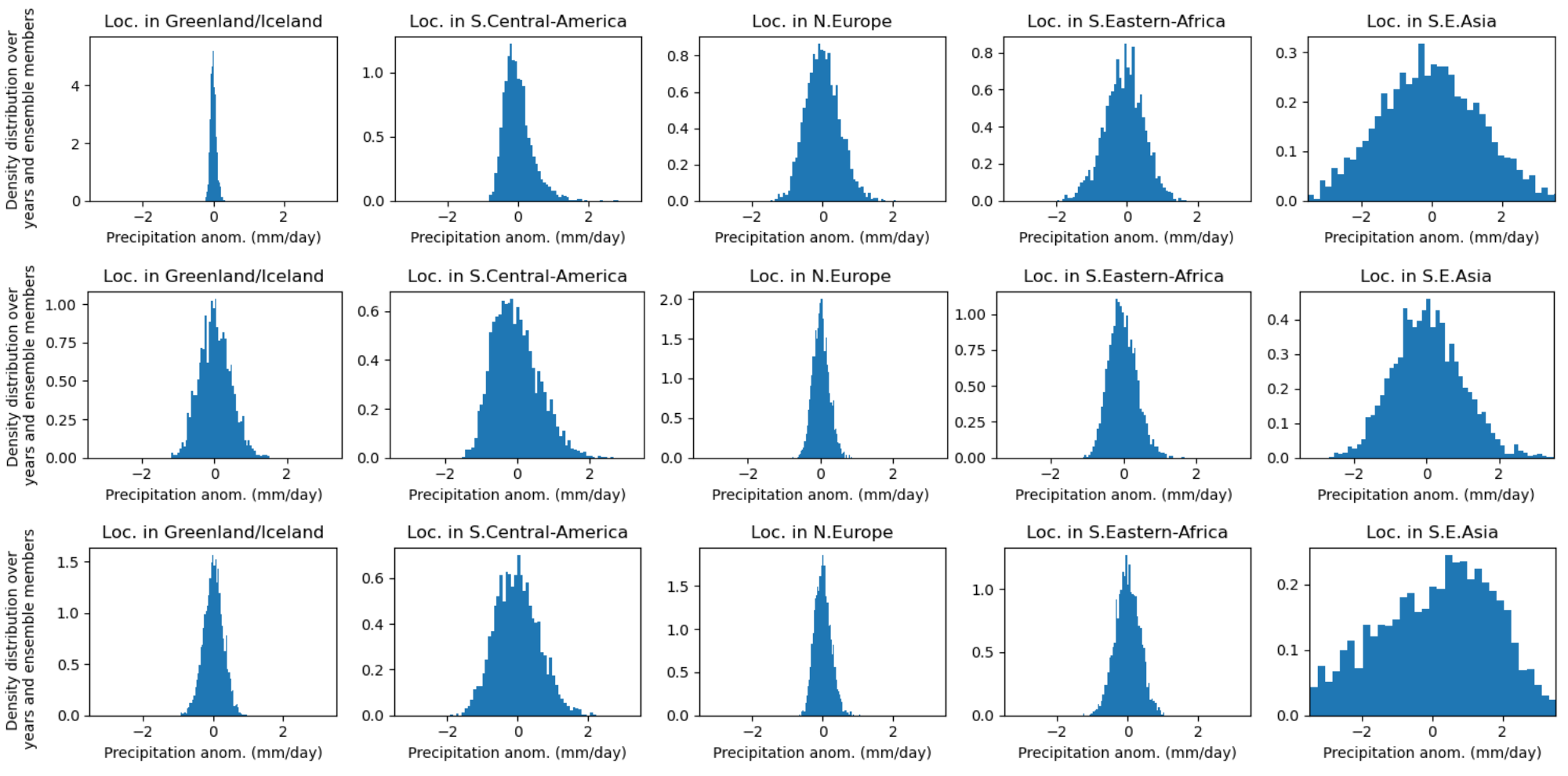}
    \label{fig:mpi_distribution_pr_over_yr_member_and_loc_in_region}
  }
\caption[]{\textbf{Distributions of precipitation anomalies in the Em-MPI dataset.} The y-axis value is the count in each bin, divided by the total number of counts and bin width, such that the density integrates to one. The x-axis is scaled to fixed values to allow for comparison between subplots.}
\label{fig:mpi_distribution_pr}
\end{figure}

\blue{The precipitation targets in the Em-MPI dataset do not follow a log-normal distribution as visualized in~\cref{fig:mpi_distribution_pr}. Instead,~\cref{fig:mpi_distribution_pr_over_loc_and_year} shows that the spatial distribution of annually- and ensemble-averaged precipitation anomalies follow a bimodal distribution that contains negative values. The mode at zero indicates locations with no projected precipitation change. With increasing $\mathrm{CO}_2^\text{cum}$  from historical over ssp126 to ssp585, the distribution's second mode increases and the overall distribution broadens towards heavier tails.~\Cref{fig:mpi_distribution_pr_over_yr_member_and_loc_in_region} shows the ensemble distributions of annually-averaged precipitation anomalies over a subset of the historical period for multiple randomly picked locations within IPCC AR6 regions. Some distributions appear to have the shape of log-normal distributions, but all distributions contain negative values. 
}

\subsection{Details of the CNN-LSTM-based emulator}\label{app:cnn-lstm}

The CNN-LSTM in the ClimateBench results table~\citep{watsonparris21climatebench}  uses the spatial input fields of $\mathrm{CO}_2$, $\mathrm{CH}_4$, $\mathrm{BC}$, and $\mathrm{SO}_2$ over a 10-year time window as input, resulting in the input shape: $(4, I, J, h_t)$ with the time history length $h_t=10$ years. The input grid size, ($I=96, J=144$), is independent of which targets are being predicted. As $\mathrm{CO}_2$ and $\mathrm{CH}_4$ are global variables, their input fields are filled with a single scalar value that is repeated along the $I,J$ dimensions. 

The emulator predicts the spatial field for the target year at the end of the time history with the output shape $(1, I, J, 1)$ with $(I,J)=(96,144) $ or $ (I^{\text{em-mpi}},J^{\text{em-mpi}})=(96,192)$ for the NorESM2-LM or Em-MPI targets, respectively. For each target climate variable, a different CNN-LSTM is trained independently. The input variables are normalized to zero-mean, unit-variance; the target variables are not normalized, following~\citep{watsonparris21climatebench}. The values for the CNN-LSTM (reproduced) are originally reported in~\citep{nguyen23climax}.

We reimplement the same CNN-LSTM as in~\citep{watsonparris21climatebench} for our internal variability experiment. The model architecture is akin to an autoencoder-LSTM-decoder structure, because the LSTM does not maintain any spatial dimensions and there are no skip connections from en- to decoder. The emulator first applies the same convolutional layer with 20 filters of size (3, 3) to every year in the input window independently. Then, the spatial dimension is reduced to 1 with a 2x2 average pooling layer and a consecutive global average pooling layer. The latent time-series with 20 feature and 1 spatial dimension(s) is fed into an LSTM layer with 25 output features and ReLU activation. Finally, a dense layer projects the features of shape (25, $h_t$) onto the output shape. Most learnable weights ($\sim 95\%$) are in this final dense output layer.

The CNN-LSTM is optimized with an objective function that minimizes the pixel- and year-wise MSE:

\begin{equation}
    \E_{t,e\sim \sD_{t,e}}\left[\frac{1}{IJ}\sum_{ij}(\hat y_{i,j,t,e} - \frac{1}{\lvert\sM\rvert} \sum_m y_{i,j,t,m,e})^2\right]
\label{eq:cnn_lstm_loss}
\end{equation}

where $\sD_{t,e} = \sT_{train} \cup \sS_{e,train}$ is the set of all years, $\sT_{train} = \{1850,1851,...,2100\}$, and scenarios, $\sS_{e,train}$, used during training. The optimizer is RMSprop with the learning rate $0.001$, no weight decay, and batch size $16$. \strikeolive{~\citep{watsonparris21climatebench} does not optimize the hyperparameters via a sweep; thus, we use the same hyperparameters for both NorESM2-LM and Em-MPI target datasets.} \blue{These hyperparameter choices are the same as in~\citep{watsonparris21climatebench}.} 
We implemented the emulator using keras 3.0 with a pytorch backend, differing from the keras 2.0+tensorflow backend in~\citep{watsonparris21climatebench}. On the Em-MPI targets dataset, the emulator has more parameters than on the ClimateBench targets dataset due to the higher resolution target grid (see~\cref{tab:em_mpi_scoreboard,tab:climatebench_pattern_scaling} for the exact number). 

\blue{It is likely possible to improve the scores of this CNN-LSTM. For example, the variance over weight initialization random seeds indicates suboptimal optimization parameters. A possible solution would be to create a separate validation set by withholding samples from the training set, and to use the validation set for tuning optimization parameters, such as learning rate and weight decay. 
But, the aim of this paper is not to develop a high-scoring emulation technique and, instead, to discuss the influence of internal variability onto benchmarking scores. 
To make a fair comparison of the role played by internal variability in the NorESM data in ClimateBench by using the Em-MPI data, we decided to follow ClimateBench's implementation of this CNN-LSTM as exactly as possible. 
Given that the CNN-LSTM implementation in ClimateBench does not use a validation set and does not optimize the hyperparameters, we also chose not to do so. 
To report statistically reliable statistics despite the variance in random weight initialization seed, we deviate from the original implementation by averaging over multiple training runs. 
}

\strikeolive{During the internal variability experiment we use the test set for early stopping instead of the validation set. As a result, the CNN-LSTM is likely less overfitted than if we would have used the validation set (which is the same as the train set). This mistake in our implementation does not affect the validity of our conclusion that the CNN-LSTM was prone to overfitting for low number of realizations. We did not rerun the experiment after noticing the mistake, because of the computational cost involved.}

\subsection{Appendix to Testing the influence of internal variability}\label{app:internal_var_exp}
The mean RMSE in the internal variability experiment for the CNN-LSTM with random weight initialization seed, $l$, is calculated as:

\begin{equation}
    \text{RMSE}_{*,n}^{exp} = \frac{1}{K}\sum_{k\in\{1,...,K\}} \frac{1}{L} \sum_{l\in\{1,...,L\}} \text{RMSE}_{*}\left(\hat f_{n,k,l}(\tX_\text{test}), 
    \mathbb{E}_{\sM} [\tY_{\text{test}}]\right)
\label{eq:rmse_cnn}
\end{equation}

\section{Extended Results \& Discussion}\label{app:results}
\subsection{Appendix to Evaluation of Linear Pattern Scaling on ClimateBench}\label{app:climatebench_results}
\label{app:linear_pattern_scaling_tas_ssp245} 
\begin{figure}[ht]
  \centering     
  \subfloat{
    \includegraphics[width=0.98\linewidth]{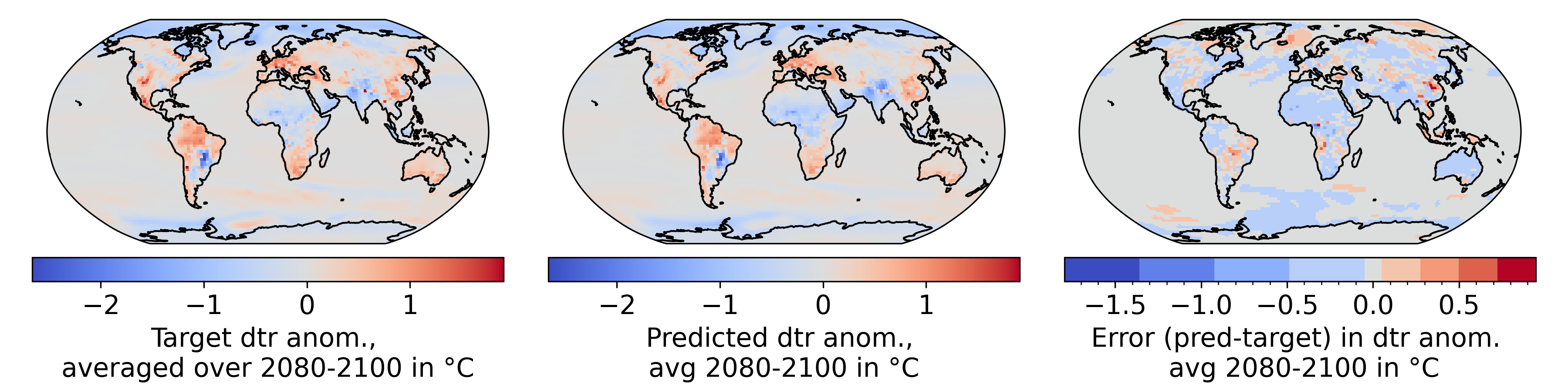}
    \vspace{-0.1in}
  }
\caption[]{\textbf{Linear pattern scaling error map for diurnal temperature range.} The left column shows the target anomalies
from the ssp245 ClimateBench test set, averaged over 3 realizations and 21 years (2080-2100). The middle
plot shows the linear pattern scaling predictions and the right plot the error of predictions minus the target.}
\label{fig:linear_pattern_scaling_error_map_diurnal_temperature_range}
\end{figure}

\begin{figure}[ht]
  \centering     
  \subfloat{
    \includegraphics[width=0.98\linewidth]{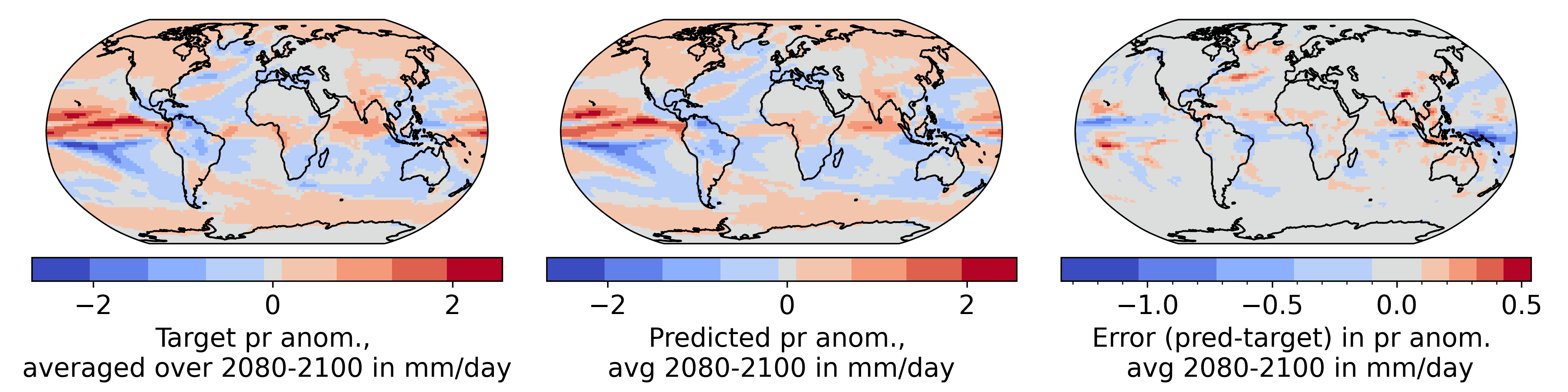}
    \vspace{-0.1in}
  }\caption[]{\textbf{Linear pattern scaling error map, similar to~\cref{fig:linear_pattern_scaling_error_map_diurnal_temperature_range}, for precipitation.}}
\label{fig:linear_pattern_scaling_error_map_precipitation}

\end{figure}

\begin{figure}[ht]
  \centering     
  \subfloat{
    \includegraphics[width=0.98\linewidth]{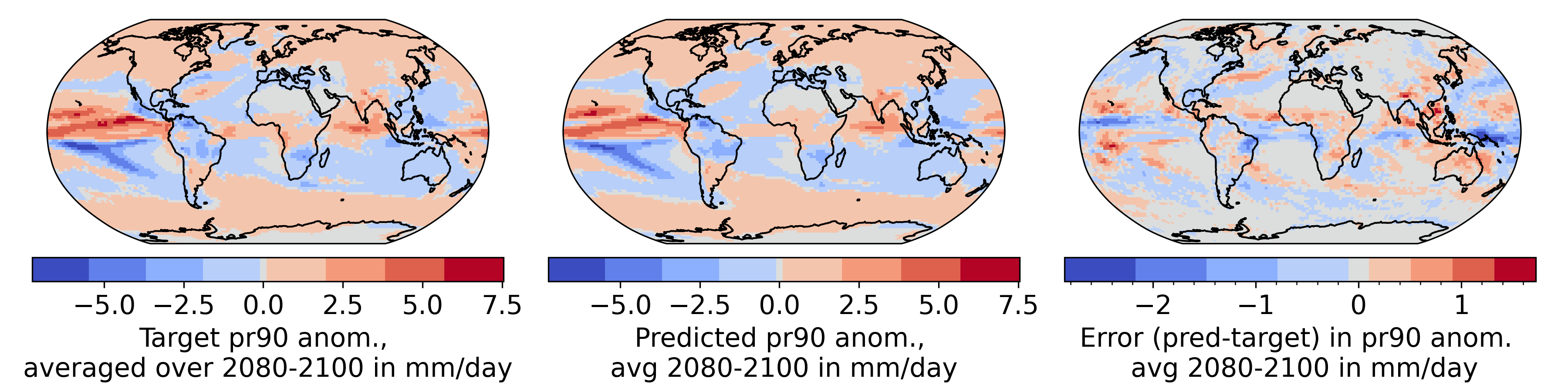}
    \vspace{-0.1in}
  }
\caption[]{\textbf{Linear pattern scaling error map, similar to~\cref{fig:linear_pattern_scaling_error_map_diurnal_temperature_range}, for 90th percentile precipitation.}}
\label{fig:linear_pattern_scaling_error_map_90th_precipitation}
\end{figure}

For completeness, we repeat the linear pattern scaling error map for diurnal temperature range, precipitation, and 90th percentile precipitation in~\cref{fig:linear_pattern_scaling_error_map_diurnal_temperature_range,fig:linear_pattern_scaling_error_map_precipitation,fig:linear_pattern_scaling_error_map_90th_precipitation}, respectively. Equivalent plots for the Gaussian Process, Neural Network, and Random Forest emulators can be found in~\citep{watsonparris21climatebench}-Fig. 4. 


\subsection{Appendix to Magnitude of internal variability in 3-member NorESM2-LM and 50-member MPI-ESM1.2-LR ensemble average}
~\Cref{fig:internal_variability_selected_regions} shows the internal variability in the 3-member ensemble mean of the NorESM2-LM (used in ClimateBench) and 50-member Em-MPI data for multiple regions. The regions were selected to \strikeolive{representatively} illustrate the local variation in the dataset.

\begin{figure}[!b]
    \centering
  \subfloat[Surface Temperature\\(continued on next page)]{
    \includegraphics[width=0.45\linewidth, trim={0 38.1cm 0 0}, clip]{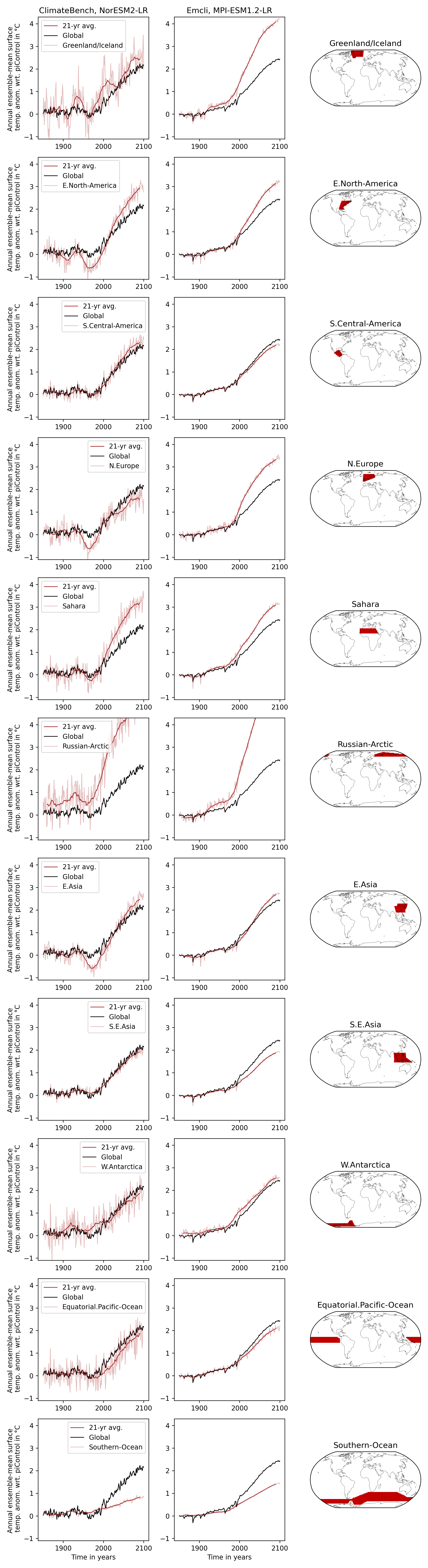}
\label{fig:internal_variability_tas_selected_regions_a}

  }\hspace{0.2in}
  \subfloat[Precipitation\\(continued on next page)]{
    \includegraphics[width=0.45\linewidth, trim={0 38.1cm 0 0}, clip]{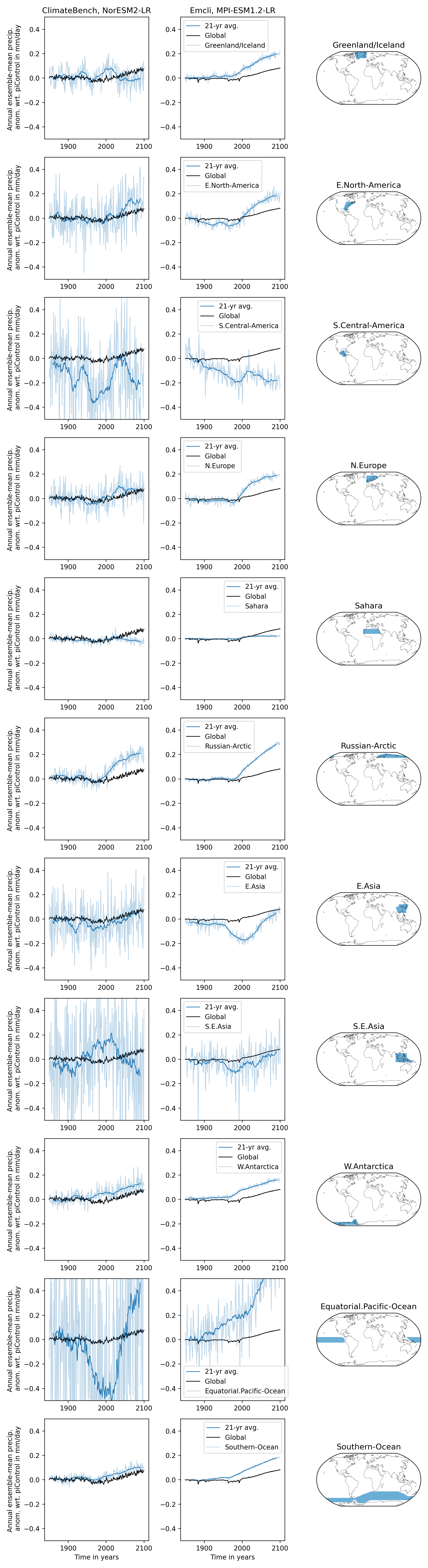}
\label{fig:internal_variability_pr_selected_regions_a}
  }
\end{figure}%
\begin{figure}[ht]
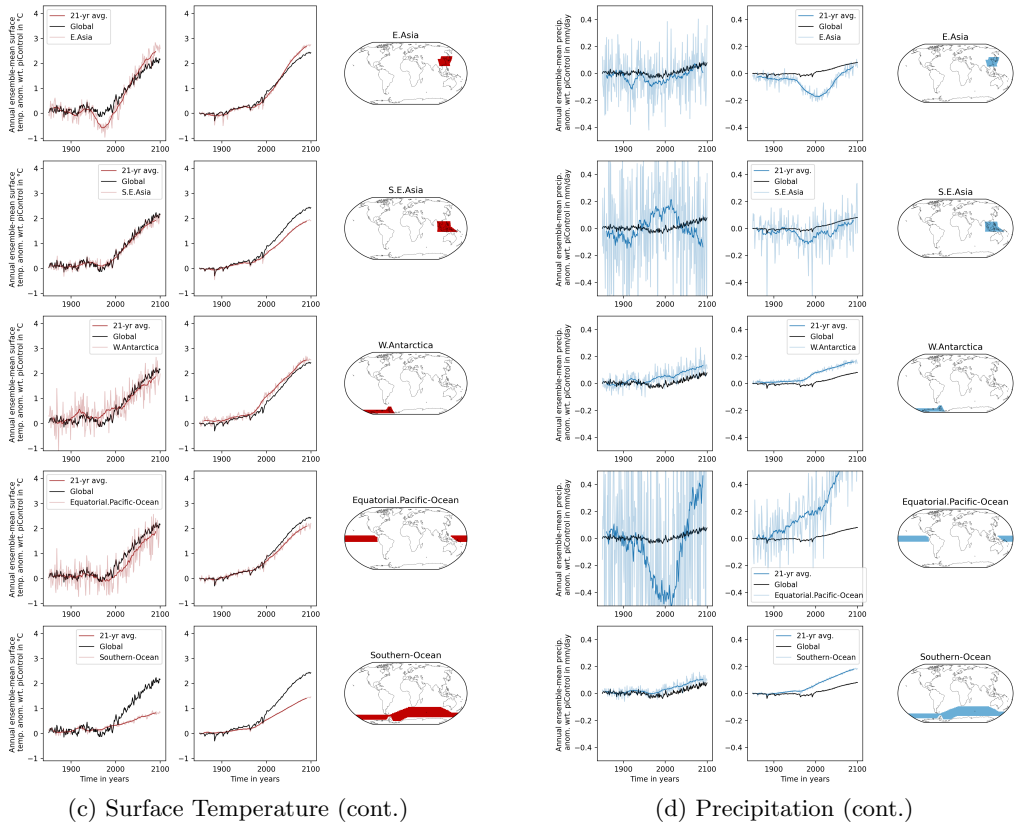
\ContinuedFloat
    \centering
  \subfloat[Surface Temperature (cont.)]{
    \includegraphics[width=0.45\linewidth, trim={0 0 0 45.5cm}, clip]{figures/mpi-esm1-2-lr/tas/internal_variability/regional_21yr_avg_selected_regions}
\label{fig:internal_variability_tas_selected_regions}
  }\hspace{0.2in}
  \subfloat[Precipitation (cont.)]{
    \includegraphics[width=0.45\linewidth, trim={0 0 0 45.5cm}, clip]{figures/mpi-esm1-2-lr/pr/internal_variability/regional_21yr_avg_selected_regions}
\label{fig:internal_variability_pr_selected_regions}
  }
    \caption[Internal variability for selected regions]{Ensemble-mean of surface temperature (a, c) and precipitation (b,d) of 3-member NorESM2-LM (left sub-columns) vs. 50-member MPI-ESM1.2-LR data (middle sub-columns) for the ssp245 scenario and selected regions (right sub-columns) representing distinct distributions. The regional averages are further averaged over each year (shaded) or 21-year window (opaque). PDF reader with zoom required.}
    \label{fig:internal_variability_selected_regions}
\end{figure}

\subsection{Appendix to Effect of internal variability on benchmarking scores}
\begin{figure}[ht]
  \centering     
  \subfloat[Spatial surface temperature in \blue{\textdegree C}]{
\label{fig:spatial_rmse_tas_inset_over_realizations}
     \includegraphics[width=0.48\linewidth, clip]{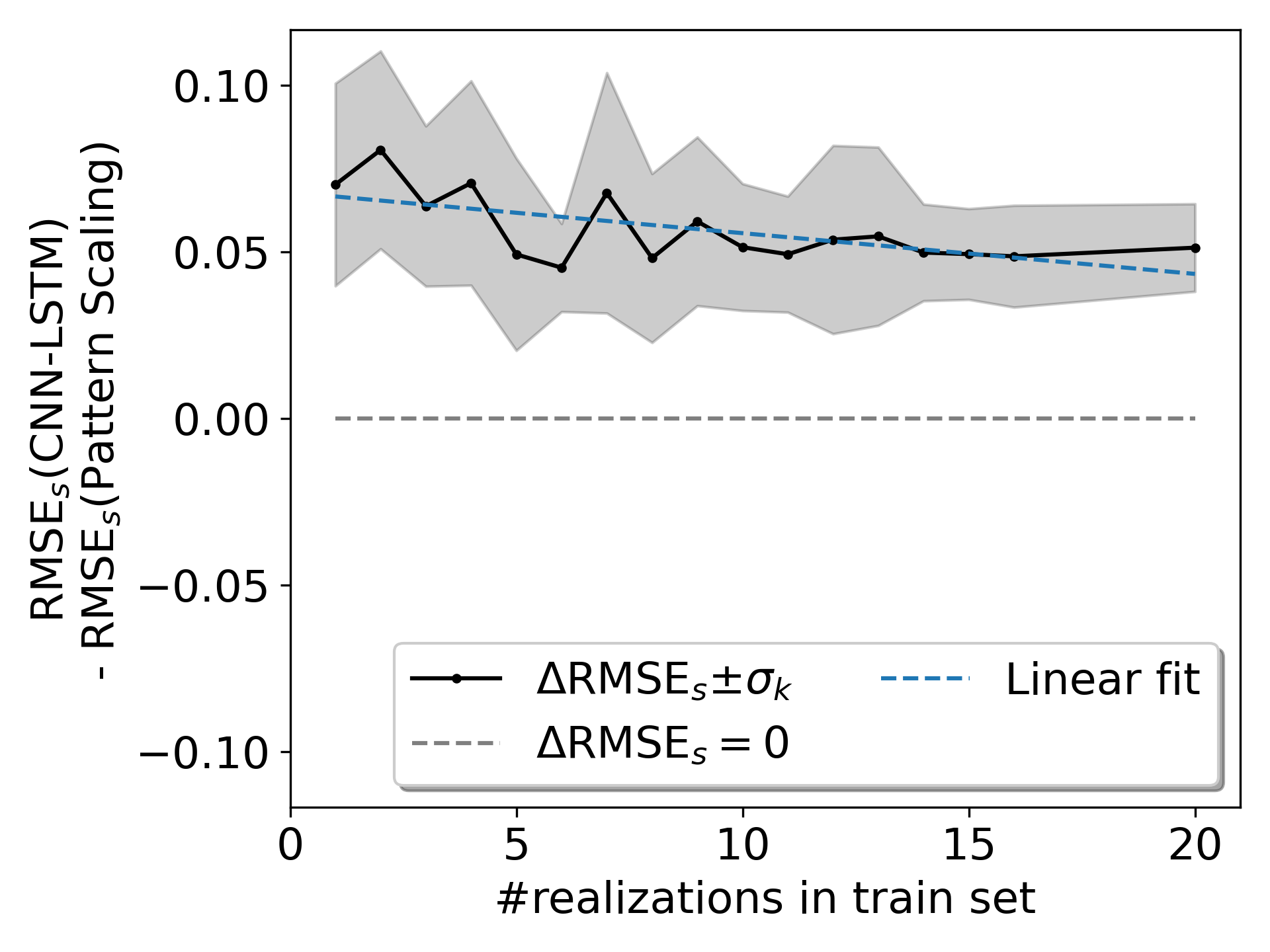}
  }
  \subfloat[Spatial precipitation \blue{in mm/day}]{
\label{fig:spatial_rmse_pr_inset_over_realizations}
     \includegraphics[width=0.48\linewidth, clip]{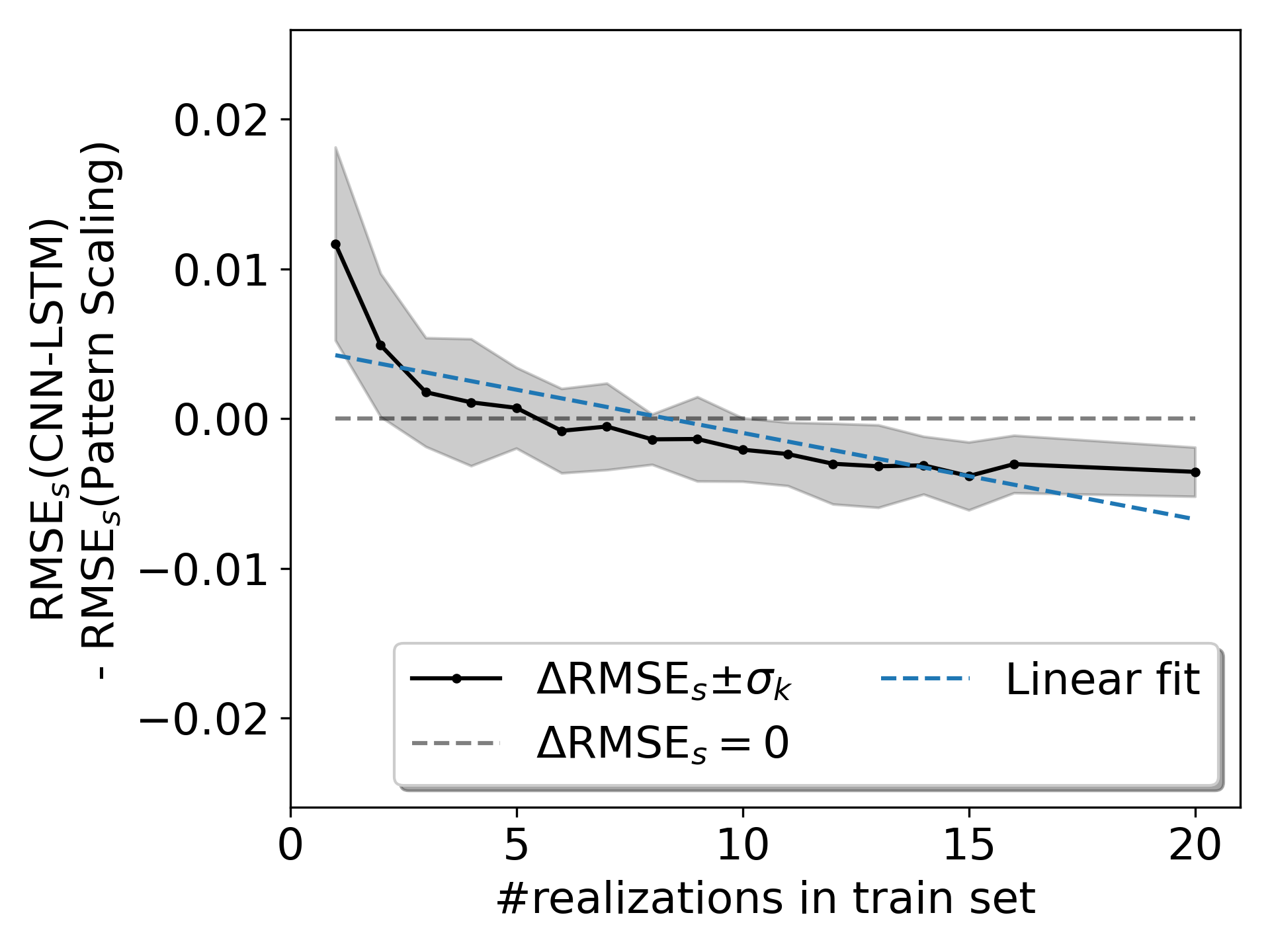}
  }
\caption[]{\textbf{Linear trend of $\Delta RMSE_{*,s}(n)$} over an x-axis inset. The plots show $\Delta RMSE$ for spatial surface temperature (left) and spatial precipitation (right) for an x-axis inset from $[0,20]$ and the best linear fit (blue) over the values in this inset. We choose a linear x-scale (instead of the log-scale in~\cref{fig:spatial_pr_over_realizations}) to illustrate the negative linear slope.}
\label{fig:spatial_rmse_inset_over_realizations}
\end{figure}

~\Cref{fig:spatial_rmse_inset_over_realizations} shows the trend of the difference in spatial RMSE, $\Delta RMSE_{*,s}(n)$ for spatial precipitation and surface temperature over an increasing number of realizations in the train set. Both climate variables show a downward trend over the x-axis inset from [0,20].

\subsection{Functional relationships of Em-MPI temperature and precipitation for multiple regions}\label{app:results_linearity}

\begin{figure}[ht!]
  \centering     
  \subfloat[Local over global surface temperature anomalies]{
    \includegraphics[width=0.98\linewidth]{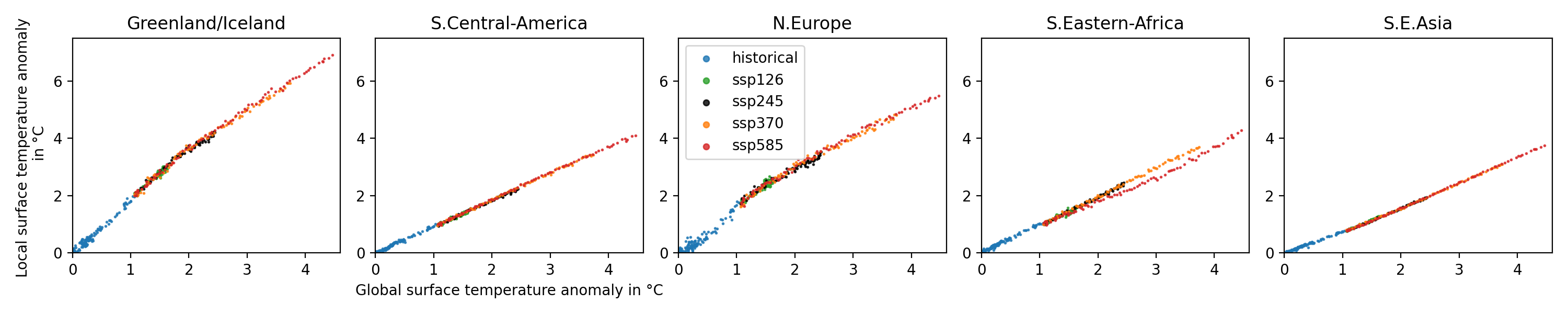}
  }\\
  \subfloat[Local precipitation over global surface temperature anomalies]{
    \includegraphics[width=0.98\linewidth]{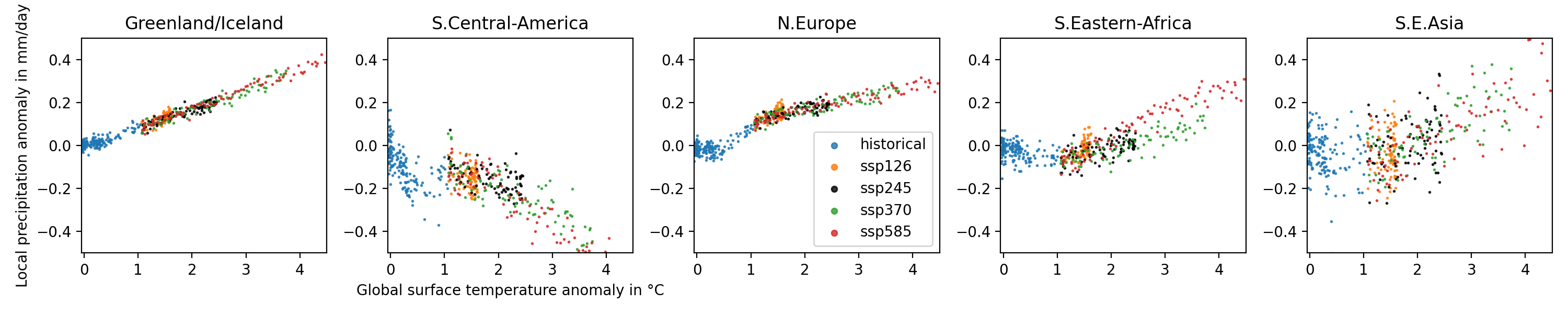}
  }
\caption[Internal variability cartoon]{\textbf{Functional relationships for multiple regions.} The top row plots regionally-averaged over globally-averaged surface temperature anomalies for multiple IPCC AR6 regions. Similarly, the bottom rows plots the regionally-averaged precipitation anomalies over global surface temperature anomalies.}
\label{fig:linearity_for_many_regions}
\end{figure}

To understand the relationships between cumulative $\mathrm{CO}_2$ emissions, surface temperature, and precipitation further, we plot the regionally-averaged values for the Em-MPI data in~\cref{fig:linearity_for_many_regions}. We selected five regions from the IPCC AR6 regions to illustrate that there are linear and nonlinear relationships depending on the location. We selected S.Eastern-Africa to highlight the nonlinear relationships in temperature and S.E.Asia for precipitation. The scatter in the precipitation plots suggests that the variable depends on more variables than global cumulative $\mathrm{CO}_2$ emissions, which goes in line with aerosols possibly influencing the precipitation in this area.

\subsection{Discussion of regional error distribution, global precipitation, and diurnal temperature range scores on ClimateBench}\label{app:climatebench_pattern_scaling}
The distribution of linear pattern scaling errors in~\cref{fig:linear_pattern_scaling_tas_ssp245} (right) shows that LPS overpredicts warming in the North Atlantic "cold blob", Weddell Sea, and Northern Asia. The emulator underpredicts warming in the Norwegian Sea and along the South Atlantic Current. The error in Northern Asia might be due to the missing aerosol inputs. The error in the Weddell and Norwegian Sea could be due to the nonlinearities from changing sea ice extent. LPS also \blue{shows sporadically located errors of small spatial extent but noteworthy magnitude} \strikeolive{error speckles} , e.g., in Central Africa or South East Asia, which could be due to internal variability in the target climate model data.

For precipitation and 90th percentile precipitation in~\cref{tab:climatebench_pattern_scaling}, linear pattern scaling has lower spatial NRMSE while ClimaX has lower global NRMSE. This possibly relates to the varying timescales in the precipitation response: If $\mathrm{CO}_2$ is rapidly increased and temperature has not yet increased, global-mean precipitation decreases due to radiative cooling; only after temperatures have increased the global-mean precipitation settles to zero again~\citep{ogorman12precip}. Whereas, local precipitation is more related to horizontal moisture transport which scales with local temperatures that increase with long-term $\mathrm{CO}_2^{cum}$ rise~\citep{ogorman12precip}. However, \blue{this discrepancy in the ClimateBench scores could also be an artifact of the internal variability influence} \strikeolive{we are also not certain if the discrepancy in the ClimateBench scores could be caused or influenced by internal variability}. ClimaX has the lowest scores for diurnal temperature range, which could be due to the strong correlation with aerosols~\citep{watsonparris21climatebench}.

\clearpage
\blue{
\section{Bias-variance experiment}\label{app:ou_experiment}
\subsection{Experiment set-up}\label{app:ou_experiment_setup}

In this pedagogical model, the target variable is a hypothetical climate variable anomaly, $y_t$, that is comparable to precipitation and relates nonlinearly to the temperature anomaly, $T_t$, via $y_t = g(T_t)$. Both variables are local, i.e., they represent the state of a generic atmosphere-ocean column. The change of the column's temperature anomaly, $dT_t$, is simulated via a local energy balance model, similar to (Eq. 6) in~\citet{giani24patternscaling}. 
\begin{align}
dT_t = \frac{r}{C} X_t dt + \frac{\lambda}{C} T_t dt + \sigma dW_t
\label{eq:local_ebm}
\end{align}
The first term on the right-hand (rhs) side represents the radiative forcing expressed as a linear function of cumulative emissions of a hypothetical greenhouse gas, $X_t$. (In analogy to carbon dioxide, the radiative forcing is assumed to be proportional to the greenhouse gas concentration, which in turn is proportional to cumulative emissions for long-lived gases.) $C$ is the effective heat capacity of the atmosphere-ocean column, and $r$ is the local perturbation radiative forcing per unit of cumulative greenhouse gas emissions. The second rhs term represents climate feedbacks that act to bring the temperature back to equilibrium and $\lambda$ is a negative feedback parameter. The last rhs term is a noise term, not standard in the energy balance model, here included representing the internal variability generated by nonlinear processes and transport terms~\citet{hasselmann76ouprocess}. The noise is sampled from a zero mean Gaussian distribution with variance $dt$ independently at each time step, $dW_t \sim \mathcal{N}(0,dt;\omega)$. $\sigma$ sets the amplitude of the internal variability and $\omega$ sets the random seed.

We simulate equation (\ref{eq:local_ebm}) by choosing a curve of cumulative emissions that looks similar to ssp245, $X_t = X_p \text{exp}(-(t-t_P)^2/2\sigma_X)$, with the peak cumulative emissions, $X_P = 5000GtX$, the timing of the peak, $t_P=250yr$, and the shape parameter, $\sigma_X = 50yr$. We choose $r=0.0008 Wm^{-2}GtX^{-1}$. The column heat capacity is given by, $C = \rho_w c_w h$, with the density of water, $\rho_w=997 kg m^{-3}$, the specific heat capacity of water, $c_w=4184 Ws kg^{-1} K^{-1}$, and the effective column-integrated water depth, $h$. We choose the values for a low-latitude ocean column in~\citet{giani24patternscaling}: $h=150m$, $\lambda=-2 Wm^{-2}K^{-1}$. We solve the equation via explicit integration in time, i.e., $T_t = T_{t-1} + dT_t dt$, with $T_0=0K$, $dt=1yr$, $t_0=0yr$, and $t_{\text{max}}=250yr$. Lastly, we chose $g$ as an arbitrary quadratic function $g(T_t) = 0.03 (4T_t)^2$.

Analogously to the ensemble averages in the internal variability experiment, we construct many training sets with differing number of realizations from this function. Each $n$-member training set is given by $\tY_{n,k} = \{y_{t,n,k} | t\in\sT\}$ with $\sT = \{t_0,...,t_{max}-1\}$ and:
\begin{align}
y_{t,n, k} = \frac{1}{n}
  \sum_{m \in \{1,...,n\}} y_t(\omega_{m,n,k})
\label{eq:ou_targets}
\end{align}
where $n$ denotes the number of realizations in a training set, $k$ denotes the $k$th random draw of equally-sized ensembles, and $\omega_{m,n,k}$ denotes the random seed for every $m-n-k$ combination. Different from the internal variability experiment, we assume an infinite set of realizations, i.e., $\omega_{m,n,k}$ is drawn with replacement, s.t., statistics are also reliable for large $n$.

Each neural network and linear model is fitted using a scalar, $X_t$, as input and output the scalar, $\hat y_{t,n,k}$.
For each $n$-$k$ combination, the training set consists of $250$ ($t_{\text{max}}$) datapoints of the function sample, $y_{t,n,k}$. For each training set, we draw a new validation set as another sample of the function, $y_{t, n,k}$, with the same $n$. The test set, for which the MSE and Bias$^2$ are reported, is the  noise-free emission-forced signal, $\bar y_t$, given by integrating~\cref{eq:local_ebm} while omitting the $\sigma dW_t$ term. The expected scores are computed with $K=2000$. 

We choose a fully-connected neural network that has two hidden layers of 64 and 32 units and a batch norm and ReLu activation after each layer. We fix the model complexity instead of sweeping over it, because a neural net can reduce to a linear regression model, and this experiment is intended to analyze the relationship between model complexity and internal variability. The neural network is optimized using MSE as the loss function and the AdamW optimizer. The best learning rate, learning rate scheduling, and weight decay were determined on the validation sets and chosen, s.t., the training process is stable and converges independent of how many realizations are in the training sets. The hyperparameters were also tuned, s.t., the training process is only marginally influenced by the weight initialization random seed. Each neural network is trained for a maximum of $150$ epoch and we implement early stopping by using each network's checkpoint with best MSE on the corresponding validation set to compute the test set scores. Additional details can be found in the code.

\begin{figure}[t]
  \centering
  \subfloat[Cumulative emissions]{
    \includegraphics[width=0.3\linewidth]{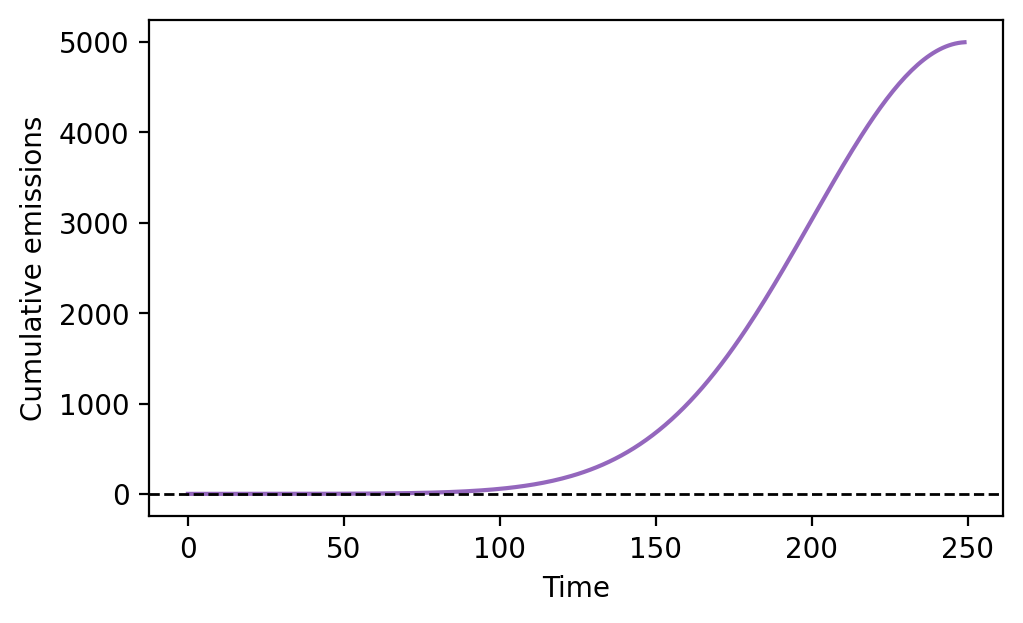}
    \label{fig:cumul_emissions}
  }  
  \subfloat[Linear fits, 2-members]{
    \includegraphics[width=0.3\linewidth]{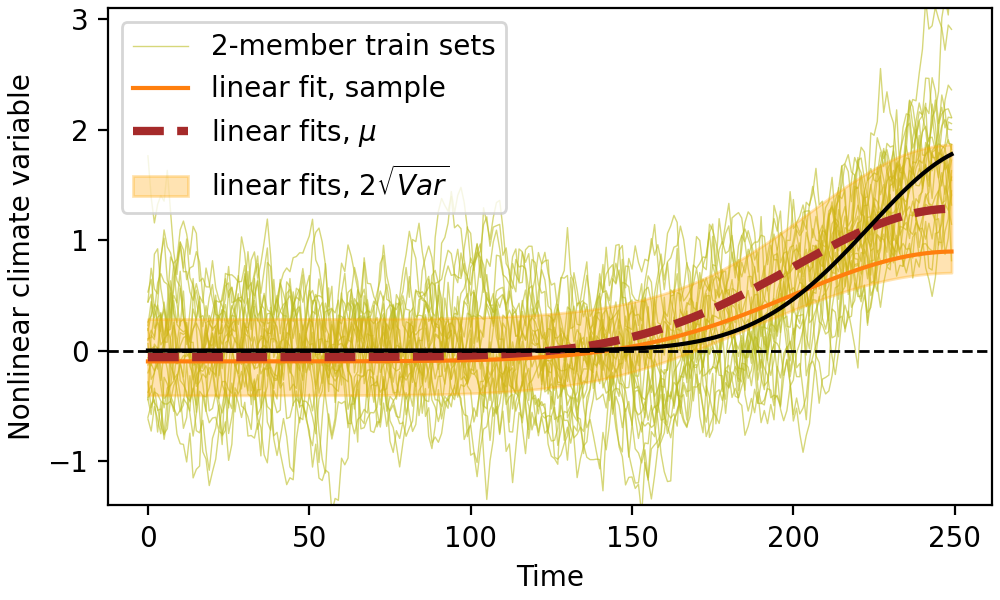}
    \label{fig:linear_fits_time_2members}
  }
  \subfloat[Linear fits, 50-members]{
    \includegraphics[width=0.3\linewidth]{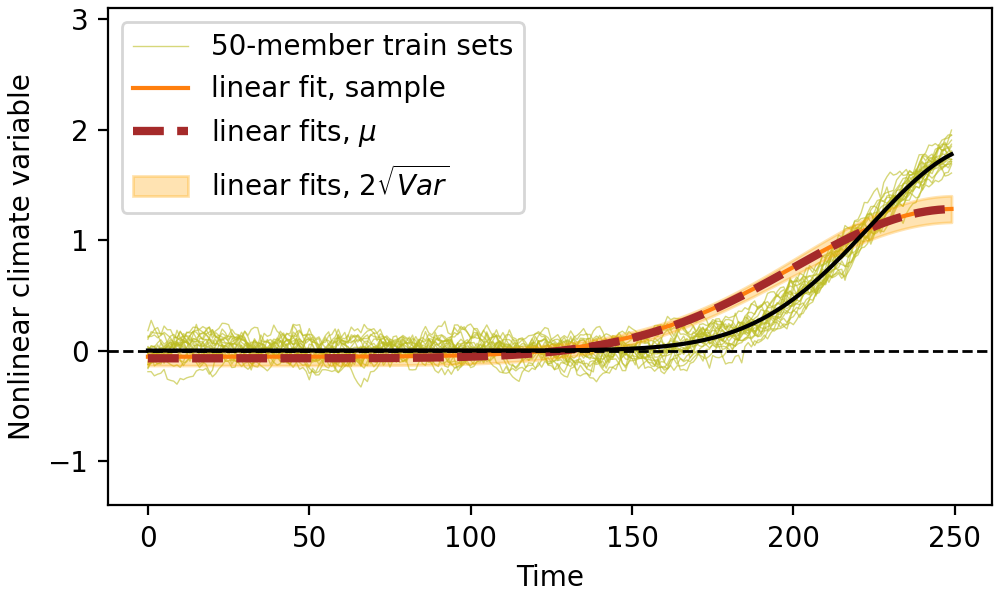}
    \label{fig:linear_fits_time_50members}
  }
  \\
  \subfloat[Nonlinear variable over time]{
    \includegraphics[width=0.3\linewidth]{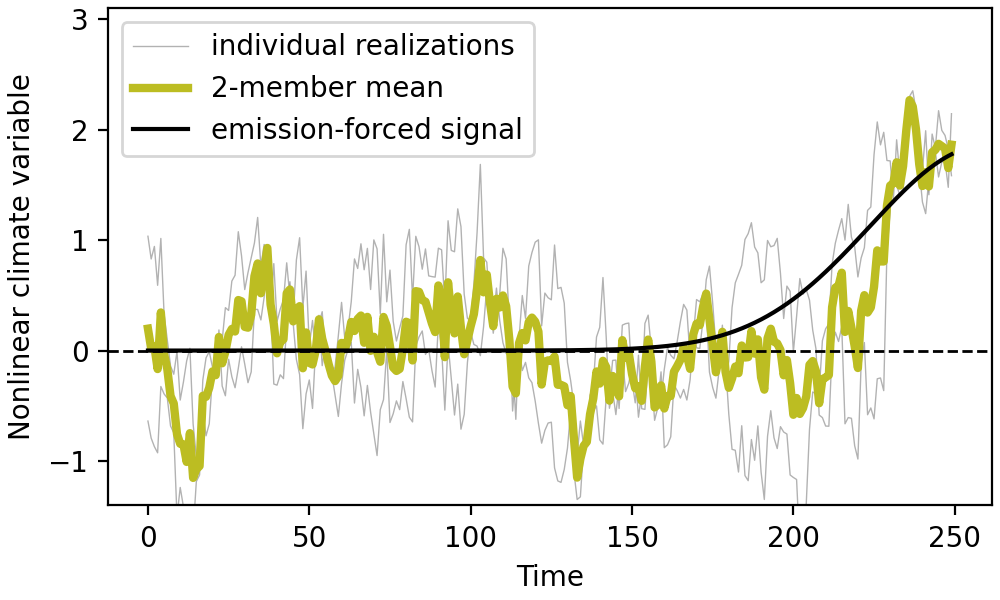}
    \label{fig:nonlinear_climate_over_time}
  }
  \subfloat[Neural fits, 2-members]{
    \includegraphics[width=0.3\linewidth]{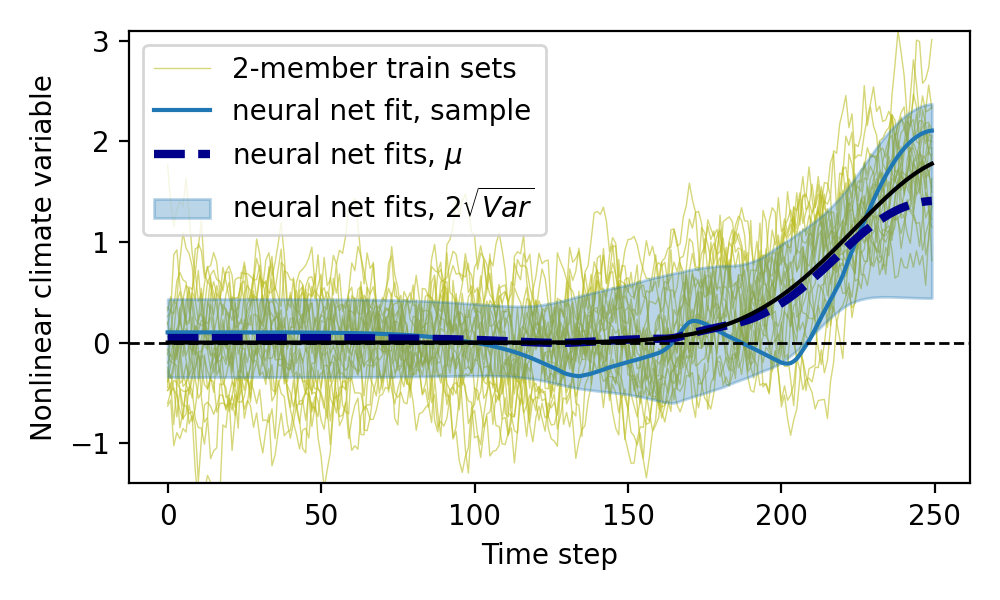}
    \label{fig:fcn_fits_2members}
  }
  \subfloat[Neural fits, 50-members]{
    \includegraphics[width=0.3\linewidth]{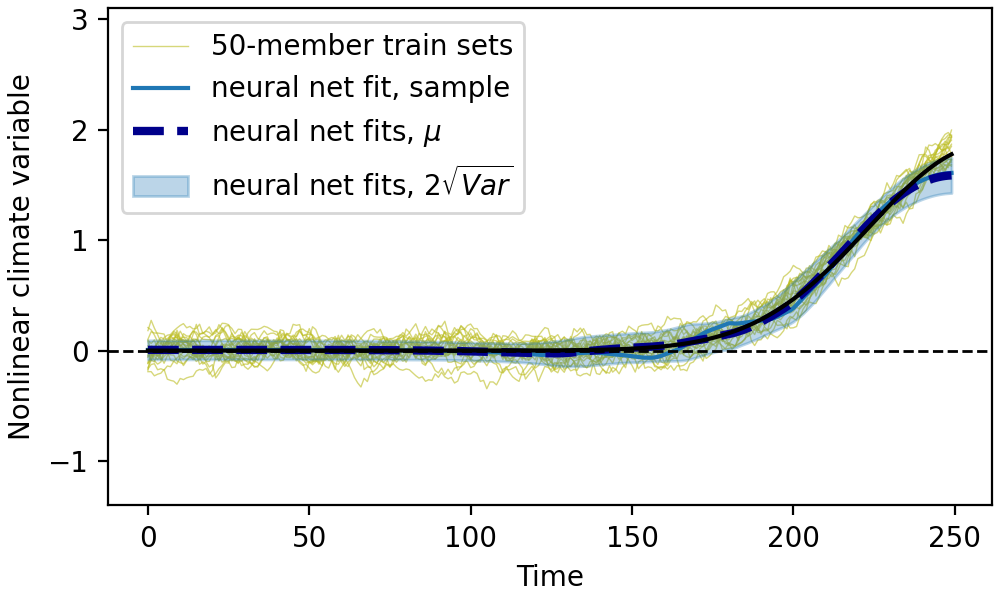}
    \label{fig:fcn_fits_time_50members}
  }\\
\caption[Details on overfitting experiment]{\textbf{Additional figures of the bias-variance experiment.}~\Cref{fig:cumul_emissions} shows the cumulative emissions of a fictitious greenhouse gas over time in GtX, which is used as input variable by the linear and neural network fit.~\Cref{fig:nonlinear_climate_over_time} shows two realizations (thin gray lines), the corresponding ensemble-mean (olive), and the emission-forced signal (black) of the nonlinear climate variable over time (instead of over cumulative emissions).~\Cref{fig:linear_fits_time_2members} shows 20x $2$-member ensemble-means (thin olive) of the nonlinear climate variable, a linear fit (orange) the sample in~\cref{fig:nonlinear_climate_over_time}, and the mean (red-dashed) and variance (orange shade) across linear fits on $2$-member training sets.~\Cref{fig:fcn_fits_2members} shows the analogous plot for neural network fits. And,~\cref{fig:linear_fits_time_50members,fig:fcn_fits_time_50members} show the analogous plots for $50$-member ensemble-mean training sets.}
\label{fig:simplified_dataset}
\end{figure}

\Cref{fig:simplified_dataset} shows additional plots illustrating the simplified dataset and the corresponding linear and neural network fits.

We compute the Fourier spectra of the signal-removed neural network fits according to:
\begin{align}
\tV_{n} = \frac{1}{K}\sum_k \lvert \text{fft}(\hat{\tY}_{n,k} - \bar{\tY})\rvert_{\text{abs}}^2
\label{eq:fourier_spectra}
\end{align}
where $\hat{\tY}_{n,k} = \{\hat y_{t,n,k} | t\in \sT\}$ is the neural network fit over time on the $k$th draw of an $n$-member ensemble-mean set and $\bar{\tY} = \{\bar{y}_t | t\in \sT\}$ is the emission-forced signal over time. We denote $\tV_{n} = \{\tV_{v,n}\ | v \in \{v_{\text{min}},...,v_{\text{max}}\}\}$ as the Fourier decomposition over frequency, $v$, and plot the decomposition over the signal period, $v^{-1}$.

\subsection{Bias-variance decomposition}\label{app:bias_variance_decomposition}

The bias-variance decomposition of the mean-squared error (MSE) can be written as follows. We compute the expected MSE of a function, $\hat f_{n,k}$, that was fit to the $k$th training subset with $n$ realizations. 
The expectation is estimated by computing it over $K$ random draws of equally-sized subsets. For the derivation, we write the MSE as a function of time, $\tau$, s.t., the expected MSE can be calculated for a selected time step or as average over a time interval:
\begin{align}
\text{MSE}_n(\tau) &= \frac{1}{K} \sum_k \left(f(\tau) - \hat f_{n,k}(\tau)\right)^2 \\
&=\mathbb{E}_K\left[\left(f(\tau) - \hat f_{n,k}(\tau)\right)^2\right] \\
&=\mathbb{E}_K\left[\left(f(\tau) - \mathbb{E}_K[\hat f_{n,k}(\tau)] + \mathbb{E}_K[\hat f_{n,k}(\tau)] - \hat f_{n,k}(\tau)\right)^2\right] \label{eq:insert_e}\\
&=\mathbb{E}_K\left[\left(f(\tau) - \mathbb{E}_K[\hat f_{n,k}(\tau)]\right)^2\right] + \label{eq:expand}\\
&\quad+ \mathbb{E}_K\left[\left(\mathbb{E}_K[\hat f_{n,k}(\tau)] -\hat f_{n,k}(\tau)\right)^2\right] + \label{eq:2nd_term}\\
&\quad+2\mathbb{E}_K\left[\left(f(\tau) - \mathbb{E}_K[\hat f_{n,k}(\tau)]\right)\left(\mathbb{E}_K[\hat f_{n,k}(\tau)] - \hat f_{n,k}(\tau)\right)\right]
\label{eq:last_term}
\end{align}
where~\cref{eq:insert_e} inserts $\pm \mathbb{E}_K[\hat f_{n,k}(\tau)]$ and \cref{eq:expand} expands the square bracket. 

The first term, \cref{eq:expand}, is the bias-squared: 
\begin{align}
\mathbb{E}_K\left[\left(f(\tau) - \mathbb{E}_K[\hat f_{n,k}(\tau)]\right)^2\right]
&=\mathbb{E}_K[f(\tau)^2] \;+ \\ &\quad+\mathbb{E}_K\left[\mathbb{E}_K[\hat f_{n,k}(\tau)]^2\right] - 2\mathbb{E}_K\left[f(\tau) \mathbb{E}_K[\hat f_{n,k}(\tau)]\right] \\
&= f(\tau)^2 + \mathbb{E}_K[\hat f_{n,k}(\tau)]^2 - 2f(\tau)\mathbb{E}_K[\hat f_{n,k}(\tau)]\\
&=\left(f(\tau) - \mathbb{E}_K[\hat f_{n,k}(\tau)]\right)^2\\
&\triangleq \text{Bias}_n(\tau)^2
\label{eq:decomposition_bias}
\end{align}
where we use the fact that $f(\tau)$ is static across all random draws, $k$, and, thus, $\mathbb{E}_K[f(\tau)] = f(\tau)$; $\triangleq$ denotes ``is equal by definition''.

The second term, \cref{eq:2nd_term}, is the variance. In particular, this variance quantifies the expected squared deviation of a model's predictions from the average of predictions from all models that were trained on randomly drawn subsets with the same number of realizations, $n$:
\begin{align}
\mathbb{E}_K\left[\left(\mathbb{E}_K[\hat f_{n,k}(\tau)] - \hat f_{n,k}(\tau)\right)^2\right] \triangleq \text{Var}_n(\tau)
\label{eq:decomposition_var}
\end{align}

We can show that the last term, \cref{eq:last_term}, is null:
\begin{align}
&\mathbb{E}_K\left[\left(f(\tau) - \mathbb{E}_K[\hat f_{n,k}(\tau)]\right)\left(\mathbb{E}_K[\hat f_{n,k}(\tau)] - \hat f_{n,k}(\tau)\right)\right] \\
=&\mathbb{E}_K\left[f(\tau)\mathbb{E}_K[\hat f_{n,k}(\tau)] - f(\tau)\hat f_{n,k}(\tau) - \mathbb{E}_K[\hat f_{n,k}(\tau)]^2 + \mathbb{E}_K[\hat f_{n,k}(\tau)]\hat f_{n,k}(\tau)\right] \\
=&f(\tau)\mathbb{E}_K[\hat f_{n,k}(\tau)] - f(\tau)\mathbb{E}_K[\hat f_{n,k}(\tau)] - \mathbb{E}_K[\hat f_{n,k}(\tau)]^2 + \mathbb{E}_K[\hat f_{n,k}(\tau)]^2 \\
=&0
\label{eq:decomposition_null}
\end{align}

In summary, the expected MSE can be decomposed into a bias-squared and variance term:
\begin{align}
  \text{MSE}_n(\tau) &=\mathbb{E}_K\left[\left(f(\tau) - \hat f_{n,k}(\tau)\right)^2\right] \\
  \dotsc&= \left(f(\tau) - \mathbb{E}_K[\hat f_{n,k}(\tau)] \right)^2 +
  \mathbb{E}_K\left[ \left(\mathbb{E}_K[ \hat f_{n,k}(\tau)]
    - \hat f_{n,k}(\tau)\right)^2\right]\\
  &=\text{Bias}_n(\tau)^2 + \text{Var}_{n}(\tau)
\label{eq:mse_decomposition}
\end{align}

The expected MSE as average over all discrete time steps, $t\in\{1,...,T\}$, similar to the global RMSE in~\cref{eq:rmse_global}, can be computed as:
\begin{align}
\text{MSE}_n &= \frac{1}{\sT}\sum_t \text{MSE}_n(t) \\
&= \frac{1}{\sT} \sum_t \text{Bias}_n(t)^2 + \frac{1}{\sT} \sum_t \text{Var}_n(t) \\
&= \text{Bias}_n^2 + \text{Var}_n
\label{eq:mse_time_avg}
\end{align}

In the bias-variance experiment, we report the expected MSE of an end-of-century 21-yr average window, similar to the spatial RMSE in~\cref{eq:rmse_spatial}. The curves, when using the global MSE, look similar (not shown) but require fewer samples to achieve statistical significance. For the 21-yr average MSE, the bias-variance decomposition can be derived in a similar way after substituting all continuous time variables with an average over the last 21 time steps, e.g., $f(\tau) \rightarrow \sum_{t\in \{T-20,...,T\}}f(t)$. 
}

\end{document}